\documentclass{article}
\usepackage{soul}
\usepackage{url}
\usepackage[hidelinks]{hyperref}
\usepackage[utf8]{inputenc}
\usepackage[small]{caption}
\usepackage{graphicx}
\usepackage{amsmath}
\usepackage{amsthm}
\usepackage{amssymb}
\usepackage{booktabs}
\usepackage{algorithm} 
\usepackage{microtype}
\usepackage{natbib}
\usepackage{caption}
\usepackage{float}
\usepackage{amsfonts}
\usepackage{nicefrac}
\usepackage{microtype}
\usepackage{mathtools}
\graphicspath{{Figures/}} 
\usepackage[labelformat=simple]{subcaption}
\usepackage[accepted]{icml2021}

\DeclarePairedDelimiter\floor{\lfloor}{\rfloor}
\newcommand\norm[1]{\left\lVert#1\right\rVert_1}
\newcommand{\R}{\mathbb{R}}

\newcommand{\varS}[1]{\sigma_{\mathcal{#1}}^2}
\newcommand{\E}{\mathop{\mathbb{E}}}
\newcommand{\intersect}{\cap}

\newtheorem{theorem}{Theorem}
\newtheorem{corollary}{Corollary}[theorem]

\begin{document}

\twocolumn[
\icmltitle{Balance is key: Private median splits yield high-utility random trees}
\icmlsetsymbol{equal}{*}

\begin{icmlauthorlist}
\icmlauthor{Shorya Consul}{a}
\icmlauthor{Sinead A. Williamson}{b}
\end{icmlauthorlist}

\icmlaffiliation{a}{Department of Electrical and Computer Engineering, The University of Texas at Austin}
\icmlaffiliation{b}{Department of Statistics and Data Science, The University of Texas at Austin}

\icmlcorrespondingauthor{Shorya Consul}{shoryaconsul@utexas.edu}
\icmlcorrespondingauthor{Sinead A. Williamson}{sinead.williamson@mccombs.utexas.edu}

\icmlkeywords{Machine Learning, ICML}

\vskip 0.3in
]

\printAffiliationsAndNotice{}

\begin{abstract}  
Random forests are a popular method for classification and regression due to their versatility. However, this flexibility can come at the cost of user privacy, since training random forests requires multiple data queries, often on small, identifiable subsets of the training data. Privatizing these queries typically comes at a high utility cost, in large part because we are privatizing queries on small subsets of the data, which are easily corrupted by added noise. 
In this paper, we propose DiPriMe forests, a novel tree-based ensemble method for differentially private regression and classification, which is appropriate for real or categorical covariates. We generate splits using a differentially private version of the median, which encourages balanced leaf nodes. By avoiding low-occupancy leaf nodes, we avoid high signal-to-noise ratios when privatizing the leaf node sufficient statistics. We show theoretically and empirically that the resulting algorithm exhibits high utility, while ensuring differential privacy.
\end{abstract}

\section{Introduction}
The prevalence of data has been one of the key drivers of technological innovation in the last decade. The abundance of data, allied with ever-increasing computing power, has driven the rapid development of sophisticated machine learning techniques, many of which have achieved hitherto unseen levels of performance. Data collection today is pervasive, across applications and devices. This has resulted in data privacy becoming a matter of public concern.

It has long been known that querying even aggregated or perturbed data can lead to leakage of private information \cite{dinur2003revealing}, motivating the development of databases and algorithms that mitigate such privacy breaches. Differential privacy \citep[DP,][]{dwork2006calibrating} is one of the most rigorous ways of analysing and ameliorating  such privacy risks. If an algorithm is $\epsilon$-DP, it means we can apply a multiplicative bound to the worst-case leakage of an individual's private information. Many algorithms have been developed with this goal in mind, such as differentially private variants of linear regression \cite{kifer2012private}, k-means clustering \cite{huang2018optimal,su2016differentially} and expectation maximization \cite{park2016dp}. 

Such privacy guarantees come at a cost---modifying an algorithm to be $\epsilon$-DP typically involves adding noise to any queries made by that algorithm, which will tend to negatively affect the algorithm's performance. This cost will tend to be higher when privatizing more complex algorithms that require multiple queries of the data, such as non-linear regression and classification algorithms \cite{SmiAlvZwiLaw2018,abadi2016deep}. One such family of non-linear regression and classification algorithms, that allows for flexible modeling but involves multiple queries, is the class of tree-based methods, such as random forests and their variants \cite{breiman2001random,geurts2006extremely}. 

Tree-based ensemble methods make minimal assumptions on the parametric forms of relationships within the data, and can be easily applied to a mixture of continuous and categorical covariates.  However, building trees that capture the appropriate structure requires many queries of the data, making them challenging to privatize. Further, since the trees partition the data into arbitrarily small subsets, the noise that must be added to each subset to ensure differential privacy can quickly swamp the signal.

We propose differentially private median (DiPriMe) forests, a novel, differentially private machine learning algorithm for nonlinear regression and classification with potentially sensitive data. 
Rather than directly privatize queries in an existing random forest framework, we start by modifying the underlying non-private trees to be robust to the addition of noise. We note that the negative impact of added privacy-ensuring noise is greatest when there are few data points associated with certain leaf nodes, especially in regression. Splitting a node based on the median value of a covariate avoids such a scenario. While using a private median, we show empirically and theoretically that private median splits lead to improved balance over alternative mechanisms. As a result, we are less likely to overwhelm the signal at each leaf node with noise.


The DiPriMe algorithm offers several advantages over existing private random forests. Because the underlying non-private algorithm is designed with privatization in mind, DiPriMe achieves impressive predictive performance across a range of tasks by minimizing the negative effects of added noise. Unlike  most existing approaches, we can easily deal with continuous or categorical covariates, without excessive additional privacy cost. Finally, we are able to support our impressive experimental results with theoretical bounds on the utility of each split---something that has not been seen in previously developed private random forests with data-dependent splits. 

We begin by reviewing the concept of differential privacy in Sec.~\ref{sec:bg}, before introducing DiPriMe forests and its variants in Sec.~\ref{sec:model}. We provide theoretical and empirical guarantees on both the privacy and the utility of our approach, and contrast our method with
existing tree-based 
methods. In Sec.~\ref{sec:experiments} we show that our method outperforms existing differentially private tree-based algorithms on a variety of classification and regression tasks.

\section{Differential Privacy}\label{sec:bg}

Differential privacy \citep[DP,][]{dwork2008differential} is a rigorous framework for limiting the amount of information that can be inferred from an individual's inclusion in a database. 
Formally, a randomized mechanism $\mathcal{F}$ satisfies $\epsilon$-DP for all datasets $D_1$ and $D_2$ differing on at most one element and all $\mathcal{S}\subseteq Range(\mathcal{F})$ if
\begin{equation}
    \mbox{Pr}[\mathcal{F}(D_1) \in \mathcal{S}] \leq e^\epsilon\mbox{Pr}[\mathcal{F}(D_2) \in \mathcal{S}]\, .
    \label{eq:dp_def}
\end{equation}

This implies that the inclusion of an individual's data can change the probability of any given outcome by at most a multiplicative factor of $e^\epsilon$. A lower value of $\epsilon$ provides a stronger privacy guarantee, as it limits the effect the omission of a data point can have on the statistic.

Typically, the mechanism $\mathcal{F}$ is a randomized form of some deterministic query $f$. To determine the degree of randomization required to satisfy \eqref{eq:dp_def}, we need to know the global sensitivity $\Delta(f) = \max_{D_1,D_2}\norm{f(D_1)-f(D_2)}$, which tells us the maximum change in the outcome of the query $f$ due to changing a single data point. Armed with the global sensitivity, we can use a number of approaches to ensure $\epsilon$-DP; we outline the two most common mechanisms below.
\paragraph{Laplace mechanism \cite{dwork2014algorithmic}} Starting with a deterministic query $f:\mathcal{D}\to \mathbb{R}^d$, where $\mathcal{D}$ is the space of possible data sets, we can construct an $\epsilon$-DP mechanism that adds appropriately scaled Laplace noise, so that   
\begin{equation}
\begin{aligned}
    \mathcal{F}(X) =& f(X) + (Y_1,Y_2,\dots,Y_d)\\ Y_i \stackrel{iid}{\sim}&\mbox{ Laplace}(0,\Delta_i(f)/\epsilon),\end{aligned}
    \label{eq:lap_mech}
\end{equation}
where $\Delta_i(f)$ is the sensitivity of the $i$-th coordinate of the output. Note how the added noise scales as $1/\epsilon$ -- increased privacy directly translates to increased noise variance.
\paragraph{Exponential mechanism \citep[EM,][]{mcsherry2007mechanism}} The Laplace mechanism assumes that our query returns values in $\mathbb{R}^d$. A more generally applicable privacy mechanism is the exponential mechanism , which allows us to pick an outcome $r\in \mathcal{R}$, where $\mathcal{R}$ is some arbitrary space. We define a scoring function $q:\mathcal{D}\times\mathcal{R} \to \R $ with global sensitivity $\Delta(q)$, and a base measure $\mu$ on $\R$. For any dataset $D \in \mathcal{D}$, selecting an outcome $r$ with probability
\begin{equation}
    \mbox{Pr}[\mathcal{F}(D)=r]\propto e^{\epsilon q(D,r)/2\Delta(q)} \times \mu(r)
    \label{eq:exp_mech}
\end{equation}
ensures $\epsilon$-DP. We shall refer to this as $\mathbf{EM(\epsilon)}$. Clearly, the scoring function $q$ should be constructed such that preferred outputs are assigned higher scores. For notational brevity, we denote $q(D,r)$ as $q(r)$ in the remainder of the paper.   

In this paper, we also use the \textbf{permute-and-flip mechanism} \citep[PFM,][]{mckenna2020permute}, a recently-proposed variant of the exponential mechanism that offers up to a two-fold improvement in the expected score of the output. Denoting the set of possible outcomes as $\mathcal{R}$, let $q_* = \max_{r\in\mathcal{R}} q(r)$ and a random permutation over $\mathcal{R}$ be $\Pi(\mathcal{R}).$ Given a privacy budget, $\epsilon$, permute-and-flip iterates over $\Pi(\mathcal{R})=\{\Pi_1, \Pi_2, \dots\}$ by drawing from $\mbox{Bernoulli}\left(\exp{\left(\frac{\epsilon}{2\Delta(q)}(q(\Pi_r)-q_*)\right)}\right)$ until a successful draw occurs and returns the corresponding outcome. Hereinafter, we shall refer to this as $\mathbf{PFM(\epsilon)}$.

Often, algorithms will involve multiple queries, requiring us to account for the overall differential privacy of the composite algorithm. We can make use of two composition theorems to obtain the overall privacy level \cite{mcsherry2009privacy}. 
\textbf{Sequential composition} tells us that a sequence of differentially private queries maintains differential privacy. Let $\mathcal{F}_i$ each provide $\epsilon_i$-DP. Then sequentially evaluating  $\mathcal{F}_i(X)$ provides $(\sum_i \epsilon_i)-$DP. 
\textbf{Parallel composition} tells us that the privacy guarantee for a set of queries on disjoint subsets of data is only limited by the worst-case privacy guarantee for any of the queries. If $\mathcal{F}_i$ each provide $\epsilon_i$-DP, and $D_i$ are arbitrary disjoint subsets of the dataset $D$, then the sequence of $\mathcal{F}_i(X\cap D_i)$ provides $(\max_i \epsilon_i)-$DP.

\section{Differentially private tree-based ensembles}\label{sec:model}
\newcommand{\RA}{\mathcal{R}_A}

Random forests \citep[RF,][]{breiman2001random} learn classification or regression rules by constructing an ensemble of decision trees, each of which outputs a collection of leaf-level parameters. 
When designing differentially private algorithms, we wish to minimize the negative impact of privatization on the overall utility. In the context of RFs, we need to privatize two steps: determining the locations of splits, and estimating the leaf node parameters. We consider options for both of these procedures in turn, before presenting our algorithm in Sec~\ref{sec:diprime}.
\paragraph{Estimating leaf node parameters}
Random forests estimate appropriate sufficient statistics at each leaf node---typically the mean in a regression setting, and the class counts in a classification context. To privatize this, we can use an $\epsilon_\ell$-DP mechanism $\mathcal{M}_\ell$ to estimate the sufficient statistics. In this paper---like almost all existing differentially private tree algorithms---we use the Laplace mechanism.
In a classification context, to render the count query at a given node $\epsilon_\ell$-DP, we can add Laplace($0, 1/\epsilon_\ell$) noise to each class count.   In the regression context, if we have $N_i$ data points associated with the $i$th leaf node, and we have a bounded target $|Y| \leq B$, we can achieve $\epsilon_\ell$-DP at a single leaf node  by adding Laplace$\left(0, 2B/N_i\epsilon_\ell\right)$ noise to the mean. Note that in both cases, the expected value of the noise added is on the order of the contribution of a single data point: changing the value of a given data point changes each class count by at most 1, and changes the sample mean by at most $2B/N_i$. Since the data points assigned to leaf nodes are disjoint, the set of all leaf-node queries is also $\epsilon_\ell$-DP, following the parallel composition theorem.

The utility of the resulting estimate at the $i$th leaf depends on two things: the value of $\epsilon_\ell$, and the number $N_i$ of data points associated with the leaf node. Small values of either $\epsilon_\ell$ or $N_i$ lead to the Laplace noise dominating the signal of interest. Therefore, to improve the utility of our estimates, we must either increase the per-leaf-node privacy budget, or increase the number of data points $N_i$. In particular, we wish to avoid the situation where the privacy budget $\epsilon_\ell$ is less than $1/N_i$, which implies that the expected magnitude of the Laplace noise is greater than that of the signal. 

\paragraph{Obtaining non-leaf node splits}
In most non-private random forests, the value at which a non-leaf node is split is determined by maximizing some score, such as the Gini index. We can privatize this by randomizing the selection procedure,  an approach taken by  \citet{friedman2010data}, \citet{patil2014differential} and \citet{fletcher2015differentially}. In these works, the authors score attribute-specific candidate splits with one child node for each category of that attribute, and then selects an attribute according to the exponential mechanism. Since each split generates one branch per distinct attribute value, any continuous attributes must be discretized.

Unfortunately, the design decisions made in these private algorithms are at odds with the goal of maximizing per-leaf node utility. In each case, a subtree is learned for each category of the chosen categorical covariate (or each unique value of the discretized continuous covariate), leading to more low-occupancy nodes than would be expected in a binary tree. This in turn leads to more low-occupancy leaf nodes where the added Laplace noise overwhelms the signal from the data, as described above. 
In addition to harming the leaf node utility, low-occupancy sub-trees are a major hindrance to good regression trees. Typically, the sensitivity of the scoring function is inversely proportional to the node count, meaning that the selection of attribute to split on will also be very noisy; sufficiently low counts would result in this selection being no better than random. 
Private versions of Extremely Random Trees \citep[ERT,][]{geurts2006extremely}, such as those proposed by \citet{bojarski2014differentially} and \citet{jagannathan2009practical}, avoid diverting budget from the leaf nodes by picking their splits entirely at random. In some cases this can improve performance, if the benefit of increased leaf node privacy budget $\epsilon_\ell$ outweighs the benefit of chasing the optimal tree structure. However, the relatively large per-leaf-node privacy budget $\epsilon_\ell$ is unfortunately paired with highly variable leaf node occupancy $N_i$, since splits are selected without considering the data distribution. 
%
\subsection{Differentially Private Median (DiPriMe) Forests}\label{sec:diprime}


\begin{algorithm}
\caption{Differentially Private Median (DiPriMe) Tree}
\label{algo:DPTree}
\begin{algorithmic}[1]
    \CLASS{\textsc{DiprimeTree}($i, i_{max}, k$)}
    \STATE  $d \leftarrow i$, $d_{max} \leftarrow i_{max}$; $K \leftarrow k$
    \ENDCLASS
    \STATE
    \FUNCTION{\textsc{FitTree}($T$, $D$, $A$, $\RA$, $B$, $\mathcal{M}_s$, $\mathcal{M}_a$, $\mathcal{M}_\ell$)}
        \IF{$T.d<T.d_{max}$}
        \STATE $a, C_a\leftarrow $
        \textsc{FindSplit}($D$,$A$,$\RA$,$B$,$\mathcal{M}_s$, $\mathcal{M}_a$,$T.K$) 
        \STATE $\RA^L,\RA^R,A_L,A_R,D_L,D_R $ \\ $\qquad \leftarrow\textsc{SplitRange}(\RA,A,a,C_a,D)$ 
        \STATE $T_R \leftarrow $ \textsc{DiprimeTree}($T.d+1$, $T.d_{max}$,$T.K$) 
        \STATE \textsc{FitTree}($T_R$,$D_R$,$A_R$,$R_{A_R}$,$B$,$\mathcal{M}_s$, $\mathcal{M}_a$, $\mathcal{M}_\ell$)
        \STATE $T_L \leftarrow $ \textsc{DiprimeTree}($T.d+1$, $T.d_{max}$,$T.K$)
        \STATE \textsc{FitTree}($T_L$,$D_L$,$A_L$,$R_{A_L}$,$B$,$\mathcal{M}_s$, $\mathcal{M}_a$, $\mathcal{M}_\ell$)
        \ELSE
        \STATE Estimate privatized mean or class counts using $\mathcal{M}_\ell$
        \STATE Store privatized mean or class counts in $T$.
        \ENDIF
    \ENDFUNCTION
    \STATE
        \FUNCTION{\textsc{FindSplit}($D$, $A$, $\RA$, $B$, $\mathcal{M}_s$, $\mathcal{M}_a$, $K$ )}
        \STATE $A_S \leftarrow$  size-$\min\{K,|A|\}$ subset of attributes $A$.
        \FORALL{$a \in A_S$}
            \IF{$a$ is categorical}
                \STATE Draw subset $C_a \subset R_a$ using $\mathcal{M}_s$ 
            \ELSE 
                    \STATE Draw location $r \in R_a$ using $\mathcal{M}_s$ 
                    \STATE $C_a = R_a \cap (-\infty, r)$
            \ENDIF
            \STATE $MSE_a \leftarrow$ mean squared error for chosen split 
        \ENDFOR
        \STATE Pick attribute $\tilde{a}$ using $\mathcal{M}_a$ (with score $MSE_a$ if $\mathcal{M}_a$ is EM or PFM) 
        \STATE \textbf{return} $\tilde{a}$, $C_{\tilde{a}}$
    \ENDFUNCTION
    \STATE
    \FUNCTION{\textsc{SplitRange}}
        \STATE {\bfseries Input: } $\RA$, $A$, $a$, $C_a$, $D$
        \STATE $\RA^L, \RA^R \leftarrow \RA$
        \STATE $A_L, A_R \leftarrow A$
        \STATE $R_a^L \leftarrow C_a, R_a^R \leftarrow R_a \setminus C_a$
        \IF{$R_a^L$ cannot be further split (single category)}
            \STATE $A_L \leftarrow A_L\setminus a$
        \ENDIF
        \IF{$R_a^R$ cannot be further split (single category)}
            \STATE $A_R \leftarrow A_R\setminus a$
        \ENDIF
        \STATE $D^L = \{(x, y) \in D: x\in \RA^L\}$
        \STATE $D^R = D \setminus D^L$
        \STATE \textbf{return} $\RA^L$,$\RA^R$,$A_L$,$A_R$,$D_L$,$D_R$    \ENDFUNCTION
\end{algorithmic}
\end{algorithm}

As discussed above, generating all possible splits for a given attribute, or selecting a split at random, can lead to low-occupancy nodes. In a non-private context, there is little downside to such behavior. But once we begin seeking differential privacy, low-occupancy nodes increase noise in the tree-selection process and lead to poor leaf-node utility. We therefore design an algorithm centered on creating balanced leaf nodes.

In a non-private context, if an attribute is continuous, we can achieve optimal leaf node balance by splitting on the median value. In a private context, we can use a differentially private estimate of the median. Let $D=(X, Y)$ be a numeric dataset with $N$ observations, to be split on attribute $a$ with bounded range $R_a = [a_L, a_U] \subset \mathbb{R}$. 
We score potential splits $r\in R_a$ according to $q(r) = -\lvert|X_{a} \intersect [a_L, r)| - |X_{a} \intersect [r, a_U]|\rvert$, noting that $q(r)$ is piecewise constant between the data points.  The sensitivity of $q(r)$ is 1, so we can achieve $\epsilon_s$-DP by selecting a split using an $\epsilon_s$-DP mechanism $\mathcal{M}_s$ (e.g.\ EM or PFM). 
If an attribute is categorical, we score all possible binary splits $(C, X\setminus C)$ as $q(C) = \lvert|C|  - |X\setminus C|\rvert$. The sensitivity of $q(C)$ is 1, so we can achieve $\epsilon_s$-DP by selecting a split $(C, X\setminus C)$ using an $\epsilon_s$-DP mechanism.

As in non-private random forests, we generate candidate splits for a size-$K$ subset of attributes, and then select between them. We score splits based on the mean squared error of the resulting partition. This scoring function has sensitivity $4B^2/N$, where the target value lies in $[-B,B]$. We then select between these $K$ candidates using an $\epsilon_s$-DP mechanism $\mathcal{M}_s$. Finally, we privatize node-level sufficient statistics using an $\epsilon_\ell$-DP mechanism $\mathcal{M}_\ell$; we choose to use the Laplace mechanism, as described earlier in this section under ``estimating leaf node parameters''.

In our analysis, we consider three versions of our algorithm. \textbf{DiPriMe} is our base algorithm: It uses EM as the median selection mechanism $\mathcal{M}_s$, and uses a uniform random variable to select an attribute in $\mathcal{M}_a$. Note that this implies that $\epsilon_a=0$, as the variable selection mechanism does not depend on the data. \textbf{DiPriMeExp} uses EM as both the median selection mechanism $\mathcal{M}_a$ and the attribute selection mechanism $\mathcal{M}_s$. \textbf{DiPriMeFlip} uses the PFM mechanism for both $\mathcal{M}_a$ and $\mathcal{M}_s$. These three choices allow us to explore the impact of data-driven attribute selection (DiPriMe vs DiPriMeExp), and the impact of our choice of private mechanism (DiPriMeExp vs DiPriMeFlip).  

As is common in tree-based algorithms, rather than using a single tree, we construct an ensemble of $N_T$ trees. We partition our data into $N_T$ subsets, and learn a DiPriMe tree on each subset. Partitioning has two benefits. First, it allows us to preserve privacy budget due to parallel composition, as discussed in the privacy analysis below.   Second, it encourages variation between the tree structures, allowing better exploration of the space. 

We summarize the process of constructing a single tree in Algorithm~\ref{algo:DPTree}. Note that the choices of $\mathcal{M}_s$, $\mathcal{M}_a$, and $\mathcal{M}_\ell$ depend on the specific model variant.
 Algorithm ~\ref{algo:DPForest} describes how we can combine multiple DiPriMe trees into a forest.  We use the following notation in Algorithms  \ref{algo:DPTree} and \ref{algo:DPForest}: $D=(X,Y)$ refers to the set of data points to which the tree $T$ is being fit. $X$ denotes the input features and $Y$, the corresponding target values. $A$ refers to the set of attributes that the tree can split on with $\RA = \{R_a: a\in A\}$ denoting the corresponding range or categories. $B$ is the upper bound on the absolute value of the target. We include code in the supplement, and will make this public upon publication.


We note that, unlike existing differentially private tree algorithms, the non-private version of our algorithm does \textit{not} correspond to a commonly used random tree model. Median splits are not typically used in random forests. Such splits are deterministic given the covariates, and agnostic to the target variable, thereby leading to higher errors. While \citet{breiman2004consistency} considers median splits for lower branches of trees as a simplifying assumption when obtaining consistency results, he categorically states that the use of median splits will result in higher error rates than greedily learned splits. Indeed, in Sec.~\ref{sec:experiments} we see that non-private median forests underperform random forests in general.

Median splits are, however, highly beneficial in a private context. Having well-balanced splits  
is critical to ensuring good performance in a differentially private setting, as we show theoretically in Sec.~\ref{sec:utility}, and empirically in both Sec.~\ref{sec:experiments} and in the supplement. 
Further, privatization of the median eliminates the deterministic nature of the median splits, allowing better performance when the optimal split is far from the median. 
\paragraph{Privacy analysis} 


We can calculate the overall privacy budget of our algorithm using the parallel and sequential composition theorems. To construct a single split, we first select a candidate split for each attribute, then select between these candidates. Since the candidate splits are on separate attributes, the total privacy cost of selecting a differentially private median for all attributes is $\epsilon_s$, making the total privacy cost of selecting private medians for each attribute, then selecting an attribute $\epsilon_s + \epsilon_a$ (where $\epsilon_a=0$ if selection is random). Since the splits at a given depth are performed on disjoint subsets of the data, the total privacy cost of building the tree is $d_{max}(\epsilon_s + \epsilon_a)$ (where $d_{max}$ is the maximum tree depth), and the total cost including privatizing the leaf nodes is $d_{max}(\epsilon_s + \epsilon_a) + \epsilon_\ell$.


\begin{algorithm}[t]
\caption{DiPriMe Forest}
\label{algo:DPForest}
\begin{algorithmic}[1]
    \CLASS{\textsc{DiprimeForest}( $n_T$, $d_{max}$, $k$, $part$)}
        \STATE $N_T \leftarrow n_T$, $\mathcal{T} \leftarrow \{\}$
        \FOR{$i=1$ {\bfseries to} $n_T$}
            \STATE $T \leftarrow$ \textsc{DiprimeTree}($0$, $d_{max},k$)
            \STATE $\mathcal{T} \leftarrow \mathcal{T}\cup T$
        \ENDFOR
    \ENDCLASS
    
    \FUNCTION{\textsc{FitForest}($F$,$D$,$A$,$\RA$,$B$,$\mathcal{M}_s$,$\mathcal{M}_a$,$\mathcal{M}_\ell$)}
        \STATE $N_T = F.N_T$
            \STATE Partition $D=(X,Y)$ into $\{D_i\}_{i=1,\dots,N_T}$
            \STATE $i \leftarrow 0$
            \FORALL{$T \in F.\mathcal{T}$}
                \STATE $T \leftarrow $ \textsc{FitTree}($T$,$D_i$,$A$,$\RA$,$B$,$\mathcal{M}_s$,$\mathcal{M}_a$,$\mathcal{M}_\ell$)
                \STATE $i \leftarrow i+1$
            \ENDFOR
    \ENDFUNCTION
\end{algorithmic}
\end{algorithm}


\subsection{Utility analysis}\label{sec:utility}

In general, the utility of a random forest---and the change in utility due to privatizing that random forest---will depend heavily on the joint distribution of the inputs $X$ and the targets $Y$. In particular, the effect of private, data-dependent split-selection mechanisms will depend on how well alternative splits capture variation in the data. This likely explains why, while utility results have been obtained for differentially private trees with random splits \cite{bojarski2014differentially}, we are not aware of any existing utility guarantees for differentially private random forests with data-dependent splits. While the utility of  DiPriMeExp and DiPriMeFlip  also depends on both $X$ and $Y$, the basic DiPriMe mechanism \textit{only} depends on the distribution of the inputs $X$, allowing us to explore the utility loss due to privatization. We start by exploring the impact of imbalanced leaf node occupancy on the utility of our estimator. In our analysis, we assume all covariates are continuous, although this can be relaxed. 

In the following theorems, we assume wlog that our dataset $D=(X,Y)$ is ordered according to some attribute $a$, such that $x_{a,1}\leq x_{a,2}\leq\cdots\leq x_{a,N}$.  Let $D_L = D_{1:\floor{N/2}}$ and $D_R = D_{\floor{N/2}+1:N}$, and let $\widetilde{D}_L = D_{1:k}, \widetilde{D}_{k+1:N}$ be the subsets obtained by a random mechanism such that $Pr\left(\min\{k,N-k\})\leq t\right)=\zeta_t$ for $0<t<N/2$.

\begin{theorem}[Regression]
\label{thm:util_bound}
 Let $SSE$ be the sum of squared errors under the split $D_L, D_R$ using exact means. Let $\widetilde{SSE}$ be the sum of squared errors under the split $\widetilde{D}_L,\widetilde{D}_R$ where the means have been rendered $\epsilon_\ell$-DP as described in Sec~\ref{sec:diprime}. Then for any $0<t<N/2$, $ \E[\widetilde{SSE}] - SSE \leq 4B^2N - 8B^2t + \frac{16B^2}{\epsilon_\ell^2 t}$ w.p.\ at least $1-\zeta_t$.
\end{theorem}

\begin{theorem}[Classification]
\label{thm:class_bound}
 Let $Acc$ be the classification accuracy under the split $D_L, D_R$ using exact counts. Let $\widetilde{Acc}$ be the accuracy under the split $\widetilde{D}_L,\widetilde{D}_R$ where the counts have been rendered $\epsilon_\ell$-DP as described in Sec~\ref{sec:diprime}. Then for any $0<t<N/2$, $\mbox{Acc} - \E[\widetilde{Acc}] \leq \left\{\frac{\epsilon_\ell t(2\theta_m-1)}{4} + \frac{1}{2}\right\}e^{-\epsilon_\ell t(2\theta_m-1)}$ 
w.p.\ at least $1-\zeta_t$ where $\theta_m$ is the minimum purity of a leaf node.
\end{theorem}

Thms~\ref{thm:util_bound} and \ref{thm:class_bound} formalize our intuition that increasing imbalance between leaf node occupancies leads to a decrease in utility. We next show that the probability of obtaining an inbalanced split under DiPriMe is upper bounded by the probability obtained under random splits.

\begin{theorem}
\label{thm:occupancy}
Consider $N$ data points in $[x_L, x_U]$, such that $R=x_U-x_L$, where the $N-2t$ centermost data points cover a range of length at least $d<R$. If we use EM to generate an $\epsilon_s$-DP median split, then the smaller of the two resulting blocks will contain at least $t$ data points w.p.\ at least $\frac{d}{Re^{-\epsilon_s}+d(1-e^{-\epsilon_s})}$.
\end{theorem}
\begin{corollary}
\label{cor:rand}
Consider $N$ data points in some bounded space. Let $N^{med}_c$ be the size of the smaller of the two resulting blocks upon splitting at the differentially private median, and $N^{rand}_c$ be the corresponding number for a randomly selected split. Then, for any $0 < t < N/2$, $Pr(N^{med}_c\leq t) \leq Pr(N^{rand}_c\leq t)$.
\end{corollary}

The above results imply a trade-off between reducing leaf-node noise by increasing $\epsilon_\ell$, and reducing leaf imbalance by increasing $\epsilon_s$. Thm~\ref{thm:occupancy} indicates that the improvement over random splits (i.e.\ $\epsilon_s=0$, corresponding to DP-ERT) will increase as the spread of the data increases. This suggests that, provided the spread of the data is sufficiently large, we will be able to obtain improved utility over DP-ERT, by expending some of our privacy budget to encourage balanced splits. We provide an empirical demonstration of Thm~\ref{thm:occupancy}, plus proofs of all theorems, in the supplement.

\subsection{Related work}\label{sec:related}
There have been several methods proposed in recent literature to learn ensembles of decision trees for classification in a differentially private manner (although, to the best of our knowledge, this paper is the first to consider regression); \citet{fletcher2019decision} provide an excellent survey. 

 Most algorithms can be seen as privatized versions of random forests \cite{breiman2001random}, which greedily choose optimal splits at internal nodes, and store sufficient statistics at the leaf nodes. Most algorithms use the Laplace mechanism to privatize leaf node counts, and differ on the method of tree construction. 
A common approach is to deterministically generate $k$-ary splits for each attribute, where $k$ is the number of unique possible values for that attribute. \citet{friedman2010data} privatizes the post-split counts using the Laplace mechanism, and then selects an attribute to split on using the exponential mechanism with entropy as the score. This method is built upon in the differentially private random forest (DP-RF) algorithm \cite{patil2014differential}, which builds multiple private trees using bootstrapped samples and considers multiple splitting metrics. 
The differentially private decision forest (DP-DF) \cite{fletcher2015differentially} uses local sensitivity to privatize the leaf-node sufficient statistics, and incorporates a pruning method. Like DiPriMe, the differentially private greedy decision forest \citet{XinYangWangHuang2019} partitions the data into disjoint subsets, to reduce the overall privacy cost. \citet{rana2015differentially} proposes a relaxation on the DP-RF approach so that the ensemble of trees preserves the variance, instead of the entire distribution of the data. This relaxation enables their method to achieve better performance, and allows for numerical covariates, at the expense of losing any claim to $\epsilon$-differential privacy.

The above algorithms all assume categorical covariates, and at each node, there is one child node per category of the selected attribute. Continuous covariates must therefore be discretized in some manner; if this is done in a data-dependent manner, we must expend privacy budget to do so \cite{kotsiantis2006discretization}. \citet{friedman2010data} do propose a method of splitting continuous covariates via the exponential mechanism, but this significantly increases the privacy budget, as discussed in \citet{fletcher2015differentially}. In either case, the use of $k$-ary splits means the resulting trees are likely to have a significant number of low-occupancy nodes when compared with DiPriMe forests. 

An alternative approach is to bypass the greedy splitting mechanism altogether, generating and selecting splits at random to give a differentially private version of the ERT algorithmsimilar to the extremely randomized trees (ERT) algorithm. \cite{jagannathan2009practical} and \cite{bojarski2014differentially} both use random splits to build their trees, allowing the entire privacy budget to be devoted to privatizing the leaf node sufficient statistics, and allowing categorical or continuous covariates. As we showed in Sec.~\ref{sec:utility}, this will lead to lower occupancy leaf nodes than DiPriMe. 

As discussed in Sec.~\ref{sec:diprime}, medians splits are not typically used in RF-type algorithms. However, they \textit{are} used in kd-trees, a tree-based technique commonly used for database querying and k-nearest-neighbor search \cite{kdtree}. Private versions of kd-trees have been proposed that make use of private medians \cite{cormode2012differentially,inan2010private}; however these works exhibit key differences from DiPriMe. While DiPriMe and other random-forest-based methods aggregate multiple weak learners to form a stronger learner, kd-trees and their private variants use a single tree, and do not aim to learn predictive rules. Further, the private kd-trees select attributes following a deterministic cycle. In a learning context this is undesirable as the ordering of attributes may unduly affect performance, and precludes the use of shallow trees if we have high-dimensional data.

In this paper, we explore two mechanisms---EM and PFM---for generating private medians. A number of alternative methods for calculating differentially private medians have been proposed \cite{nozari2017differentially,tzamos2020,bohler2020} and could be used in DiPriMe; however the comparable performances of DiPriMeExp and DiPriMeFlip suggest that the use of alternative mechanisms would have little impact on performance.

\section{Experiments}\label{sec:experiments}
\begin{table*}[ht]
    \centering
    \small
    \caption{MSEs on various regression tasks, $N_T=10,K=5,\epsilon=10,\rho=0.5$ ($d_{max}$ specified in parentheses).}
    \label{tab:reg_err}
    \vskip 0.1in
    \begin{tabular}{lccccc}
        \toprule
        & Parkinson's & Appliances & Traffic & Turbine & Flight \\
        & $(\times 10^{-2})$ & $(\times 10^{-3})$ & $(\times 10^{-3})$ & $(\times 10^{-4})$ & $(\times 10^{-4})$\\
        \midrule
         RF & $1.59\pm 0.07$ $(7)$ & $6.27\pm 0.48$ $(9)$ & $4.64\pm 0.20$ $(10)$& $1.22\pm 0.04$ $(10)$& $1.98\pm 0.06$ $(14)$\\ 
         ERT & $2.56\pm 0.16$ $(7)$ & $7.17\pm 0.50$ $(9)$& $6.31\pm 0.35$ $(10)$& $2.25\pm 0.11$ $(10)$& $2.05\pm 0.06$ $(14)$\\
         Median & $1.88\pm 0.09$ $(7)$ & $6.17\pm 0.43$ $(9)$& $6.66\pm 0.41$ $(10)$& $1.08\pm 0.04$ $(10)$& $2.09\pm 0.07$ $(14)$\\
         DP-RF & $5.61\pm 1.26$ $(3)$ & $40.7\pm 6.2$ $(4)$& $33.1\pm 5.6$ $(4)$& $241\pm 56$ $(4)$ & $407\pm 26$ $(6)$\\
         DP-ERT & $3.46\pm 0.14$ $(7)$ & $8.72\pm 0.62$ $(9)$& $43.2\pm 3.3$ $(10)$& $41.9\pm 9.9$ $(10)$& $2.72\pm 0.08$ $(14)$\\
         DiPriMe & $3.36\pm 0.13$ $(4)$ & $8.58\pm 0.63$ $(6)$& $50.6\pm 5.5$ $(7)$& $15.7\pm 0.9$ $(7)$& $2.38\pm 0.07$ $(11)$\\
         DiPriMeExp & $3.32\pm 0.10$ $(4)$ & $8.53\pm 0.60$ $(6)$& $31.8\pm 2.3$ $(7)$& $9.60\pm 1.11$ $(7)$& $2.38\pm 0.07$ $(11)$\\
         DiPriMeFlip & $3.30\pm 0.12$ $(4)$ & $8.46\pm 0.60$ $(6)$& $24.7\pm 2.9$ $(7)$& $8.12\pm 0.66$ $(7)$& $2.32\pm 0.06$ $(11)$\\
         \bottomrule
    \end{tabular}
\end{table*}
\begin{figure*}[ht]
    \centering
    \begin{subfigure}[t]{0.3\textwidth}
        \includegraphics[width=\textwidth]{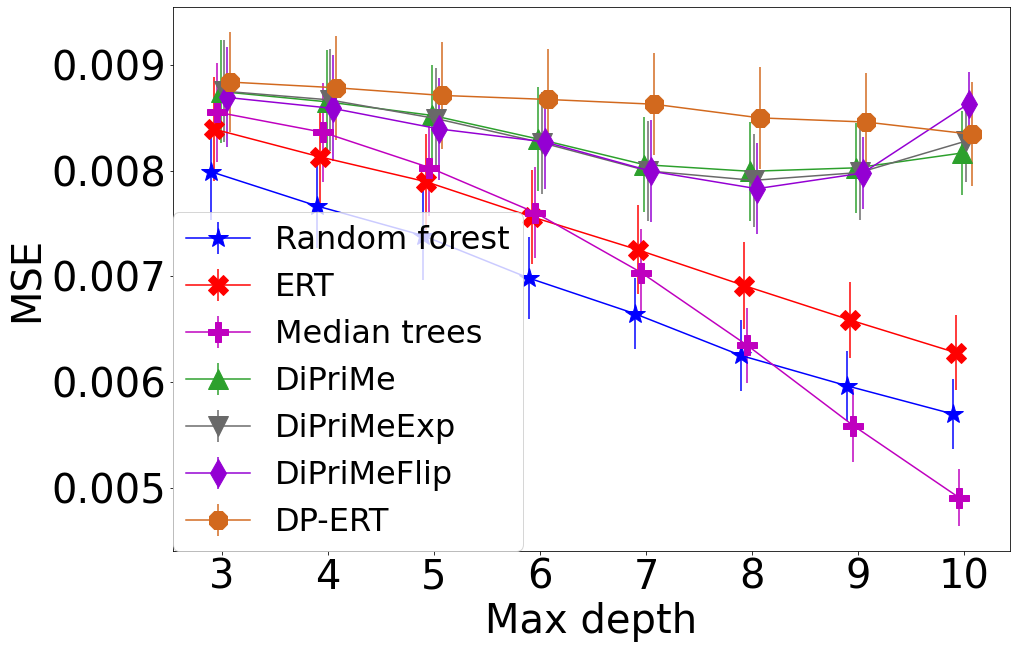}
        \caption{$\epsilon=10$, $N_T=10$}
        \label{fig:reg_maxd}
    \end{subfigure}
    \centering
    \begin{subfigure}[t]{0.3\textwidth}
        \includegraphics[width=\textwidth]{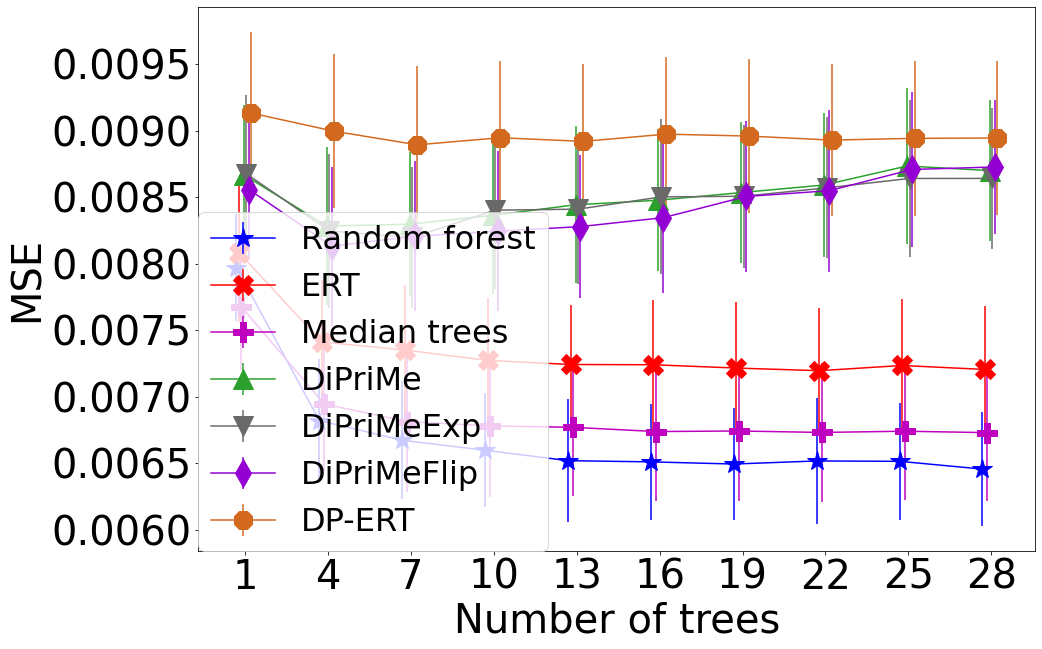}
        \caption{$\epsilon=10$, $d_{max}=8$}
        \label{fig:reg_ntree}
    \end{subfigure}
    \centering
    \begin{subfigure}[t]{0.3\textwidth}
        \includegraphics[width=\textwidth]{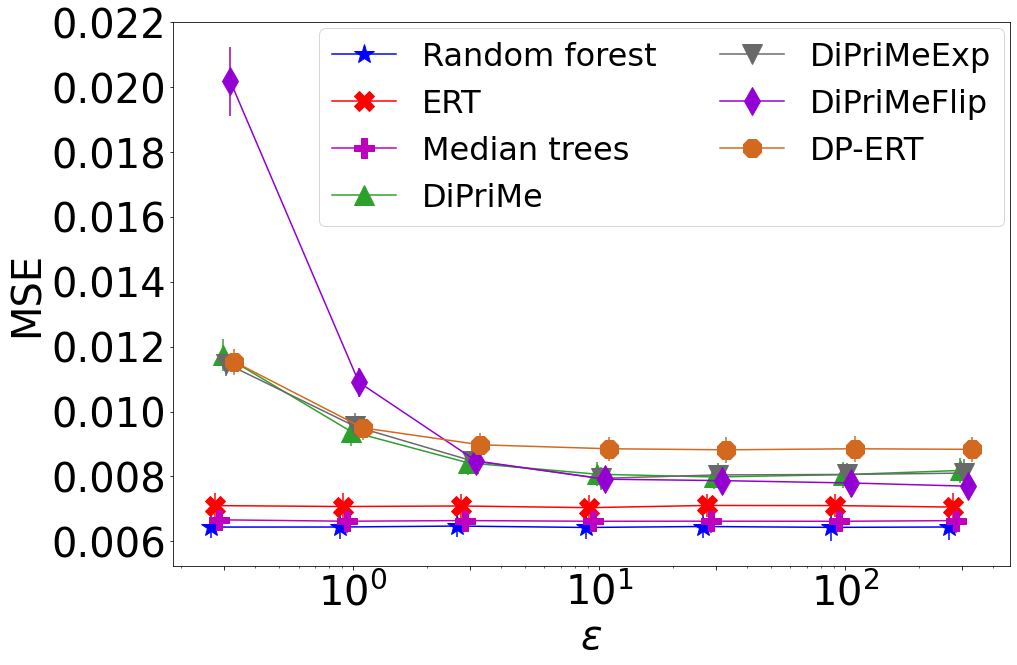}
        \caption{$d_{max}=8$, $N_T=10$}
        \label{fig:reg_eps}
    \end{subfigure}
    \caption{MSEs on the Appliances Energy dataset for various values of $\epsilon$, $d_{max}$ and $N_T$ on the Appliances Energy prediction dataset.}
    \label{fig:reg_param}
\end{figure*}
To explore the utility of DiPriMe, we look at estimate qualities obtained across a range of regression and classification tasks, comparing against state-of-the-art private and non-private algorithms. For classification, we compare against two privatized versions of random forests (DP-DF \cite{fletcher2015differentially} and DP-RF \cite{patil2014differential}), and a privatized version of ERT  \citep[DP-ERT,][]{bojarski2014differentially}). There has been little focus on regression in the literature, limiting our comparisons. 
We modified DP-ERT to estimate the mean, rather than counts, at each leaf. We also created a regression analogue of DP-RF, where splits are scored by mean squared error and means are stored instead of class counts.  The innovations introduced in DP-DF are specific to classification, precluding a simple analogue for regression.
For all methods, the number of trees, $N_T$, was set a-priori to 10 unless otherwise stated,  and  $d_{max}$ was chosen such that the expected occupancy at each leaf node was greater than some number (25 in most experiments, 10 for Banknote Authentication data). For DiPriMeExp and DiPriMeSplit, we split the overall privacy budget $\epsilon$  by setting $\epsilon_\ell = (1-\rho)\epsilon$, and $\epsilon_s = \epsilon_a = \rho \epsilon / 2 d_{max}$, for some $\rho \in (0, 1)$. For DiPriMe, we set $\epsilon_\ell = (1-\rho)\epsilon$, and $\epsilon_s =\rho \epsilon /d_{max}$. Unless otherwise stated, we used $\rho=0.5$. For all methods except DiPriMe and ERT, we selected $K=5$ candidate features for each split.
\subsection{Regression}
We consider five regression datasets: the Parkinson's telemonitoring dataset ($N=5875$), the Appliance Energy Prediction dataset ($N=19735$), Metro Interstate Traffic Volume dataset ($N=48204$) and Gas Turbine Emission dataset ($N=36733$) from the UCI Machine Learning Repository \cite{Dua:2019}, and the Flight Delay dataset used by \citet{jagannathan2009practical}. 
For the purposes of computational complexity, we sampled 800,000 data instances from the Flight Delay dataset for this experiment. We bin the numeric features into 5 bins for DP-RF as it requires categorical features. For each dataset, we scaled the target variable to lie in $[0,1]$, took 90\% of the data as the training set and computed the mean squared error (MSE) over the test set. 
We show the resulting mean squared errors in Table \ref{tab:reg_err}, where $\epsilon=10$ for the private methods. 
%
%


Table~\ref{tab:reg_err} shows that, even though median forests typically underperform RFs in a non-private context, DiPriMeExp and DiPriMeFlip clearly outperform both DP-ERT and DP-RF. We note that DiPriMe tends to perform comparably to DiPriMeExp and DiPriMeFlip in datasets where RFs and ERTs have similar MSE. This is perhaps not surprising, since comparable performance of RFs and ERTs suggest there is less to be gained in these cases from target-dependent split selection. Conversely, DiPriMe performs poorly on the Traffic dataset, where RFs perform significantly better than ERT.  Also of note is the very poor performance of DP-RF. This is likely due to the need to pre-discretize continuous covariates: too coarse discretization will lead to a poor approximation in the non-private setting, and too fine discretization will lead to low-occupancy nodes.

In Fig.~\ref{fig:reg_param}, we see how performance varies with privacy budget $\epsilon$, the number of trees in the forest $N_T$, and the maximum tree depth $d_{max}$, on the Appliances dataset. We have omitted DP-RF from these figures as it vastly underperforms the other methods. We see that for most parameter settings, all the variants of DiPriMe outperform DP-ERT.  Fig.~\ref{fig:reg_maxd} illustrates the inherent trade-off between learning deeper trees and utility. Deeper trees give a finer approximation of the data, demonstrated by the decreasing MSE of the non-private methods. However, this deteriorates the utility of DiPriMe by (a) reducing the privacy budget for the split at each node (b) increasing the sensitivity of the mean at the leaf nodes as there are likely to be fewer data instances in deeper nodes. Increasing the number of trees in the ensemble results in a similar trade-off, as shown in Fig.~\ref{fig:reg_ntree}; the number of data points to learn each tree is inversely related to the number of trees in the ensemble. So, while more trees are expected to generally reduce the mean squared error, each tree has less data to learn from. The degradation in performance of DiPriMeFlip vis-\`{a}-vis DiPriMeExp and DiPriMe at low values of $\epsilon$ (see Fig.~\ref{fig:reg_param}(c)) can be attributed to the fact that PFM is guaranteed to give lower error than EM only for $\epsilon>0.86$ \cite{mckenna2020permute}.

\begin{figure}[ht]
    \centering
    \includegraphics[width=0.35\textwidth]{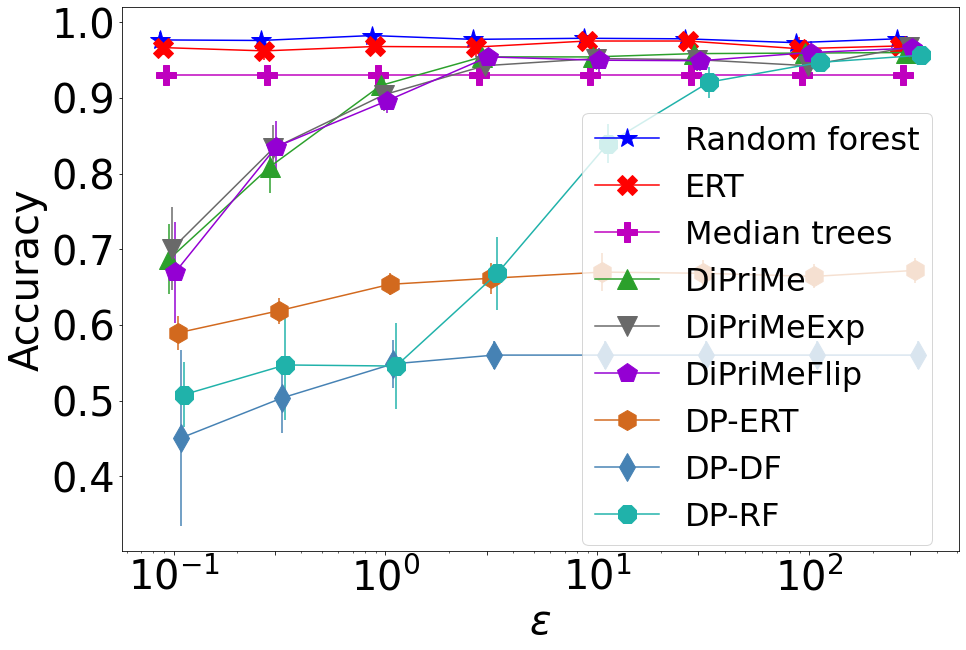}
        \caption{Accuracy on the Banknote Authentification data, for various values of $\epsilon$ ($d_{max} = 5$, $N_T=10$, $\rho = 0.5$).}
    \label{fig:class_eps}
\end{figure}

\subsection{Classification}
\begin{table*}[ht!]
    \centering
    \caption{Accuracy on classification tasks, $N_T=10, K=5, \epsilon=2,\rho=0.5$ ($d_{max}$ specified in parentheses).}
    \label{tab:class_err}
    \vskip 0.1in
    \small
    \begin{tabular}{lccccc}
        \toprule
        & Banknote & Credit Card & Robot & Higgs & PUC-Rio \\
        \midrule
        RF & $0.989\pm 0.007$ $(6)$& $0.819\pm 0.005$ $(10)$& $0.987\pm 0.004$ $(7)$& $0.702\pm 0.003$ $(11)$& $1.000\pm 0.000$ $(12)$\\
        ERT & $0.983\pm 0.009$ $(6)$& $0.814\pm 0.006$ $(10)$& $0.878\pm 0.008$ $(7)$& $0.648\pm 0.003$ $(11)$& $1.000\pm 0.000$ $(12)$\\
        Median & $0.985\pm 0.009$ $(6)$& $0.789\pm 0.005$ $(10)$& $0.953\pm 0.004$ $(7)$& $0.695\pm 0.003$ $(11)$& $1.000\pm 0.000$ $(12)$\\ 
        DP-DF & $0.557\pm 0.029$ $(2)$& $0.781\pm 0.005$ $(4)$& $0.600\pm 0.014$ $(3)$& $0.530\pm 0.003$ $(5)$& $0.515\pm 0.004$ $(5)$\\ 
        DP-RF & $0.853\pm 0.024$ $(2)$& $0.792\pm 0.005$ $(4)$& $0.728\pm 0.021$ $(3)$& $0.540\pm 0.003$ $(5)$& $0.985\pm 0.009$ $(5)$\\ 
        DP-ERT & $0.684\pm 0.022$ $(6)$& $0.674\pm 0.004$ $(10)$& $0.594\pm 0.008$ $(7)$& $0.506\pm 0.001$ $(11)$& $0.787\pm 0.022$ $(12)$\\
        DiPriMe & $0.910\pm 0.011$ $(3)$& $0.783\pm 0.006$ $(6)$& $0.759\pm 0.015$ $(4)$& $0.599\pm 0.007$ $(8)$& $0.999\pm 0.001$ $(8)$\\
        DiPriMeExp & $0.907\pm 0.021$ $(3)$& $0.784\pm 0.005$ $(6)$& $0.762\pm 0.016$ $(4)$& $0.606\pm 0.005$ $(8)$& $1.000\pm 0.000$ $(8)$\\
        DiPriMeFlip & $0.903\pm 0.012$ $(3)$& $0.781\pm 0.005$ $(6)$ & $0.784\pm 0.019$ $(4)$& $0.608\pm 0.005$ $(8)$& $1.000\pm 0.000$ $(8)$\\
        \bottomrule
    \end{tabular}
\end{table*}

We use five datasets from the UCI Machine Learning Repository \cite{Dua:2019}---the Banknote Authentication ($N=1372)$, Credit Card Default ($N=30000$), Wall-Following Robot Navigation ($N=5456$), HIGGS  and the PUC-Rio datasets($N=122244$)---to compare the performance of our DiPriMe to DP-DF, DP-RF, and DP-ERT. We use a randomly selected size-$10^5$ subset of the HIGGS data for our experiments. 
As DP-DF and DP-RF require categorical features, we bin the numeric features into 5 bins. 

Table~\ref{tab:class_err} shows the classification errors obtained by each method. In the non-private setting, the performance of median forests is consistently worse than RFs, with ERTs a little worse off. However, as we introduce privacy, we see that the performance of DiPriMe and its variants far exceed that of DP-ERT, DP-RF and DP-DF. This can likely be attributed to (a) DiPriMe's capability to directly utilize and split on numeric features without the need for prior discretization (b) the robustness of the median construction to privatization, due to its preference for balanced node splits. There is little to choose between the variants of DiPriMe. As we see in  Fig.~\ref{fig:class_eps}, while the variants of DiPriMe have a notable  loss of accuracy compared with the non-private  algorithms for small values of $\epsilon$, we get comparable performance as $\epsilon$  increases. By contrast, DP-ERT and DP-DF continue to underperform even as $\epsilon$ increases, and DP-RF requires much larger $\epsilon$ to attain comparable performance.

\subsection{Balancedness of splits} \label{sec:expt_splits}
\begin{figure}
    \centering
    \includegraphics[width=\linewidth]{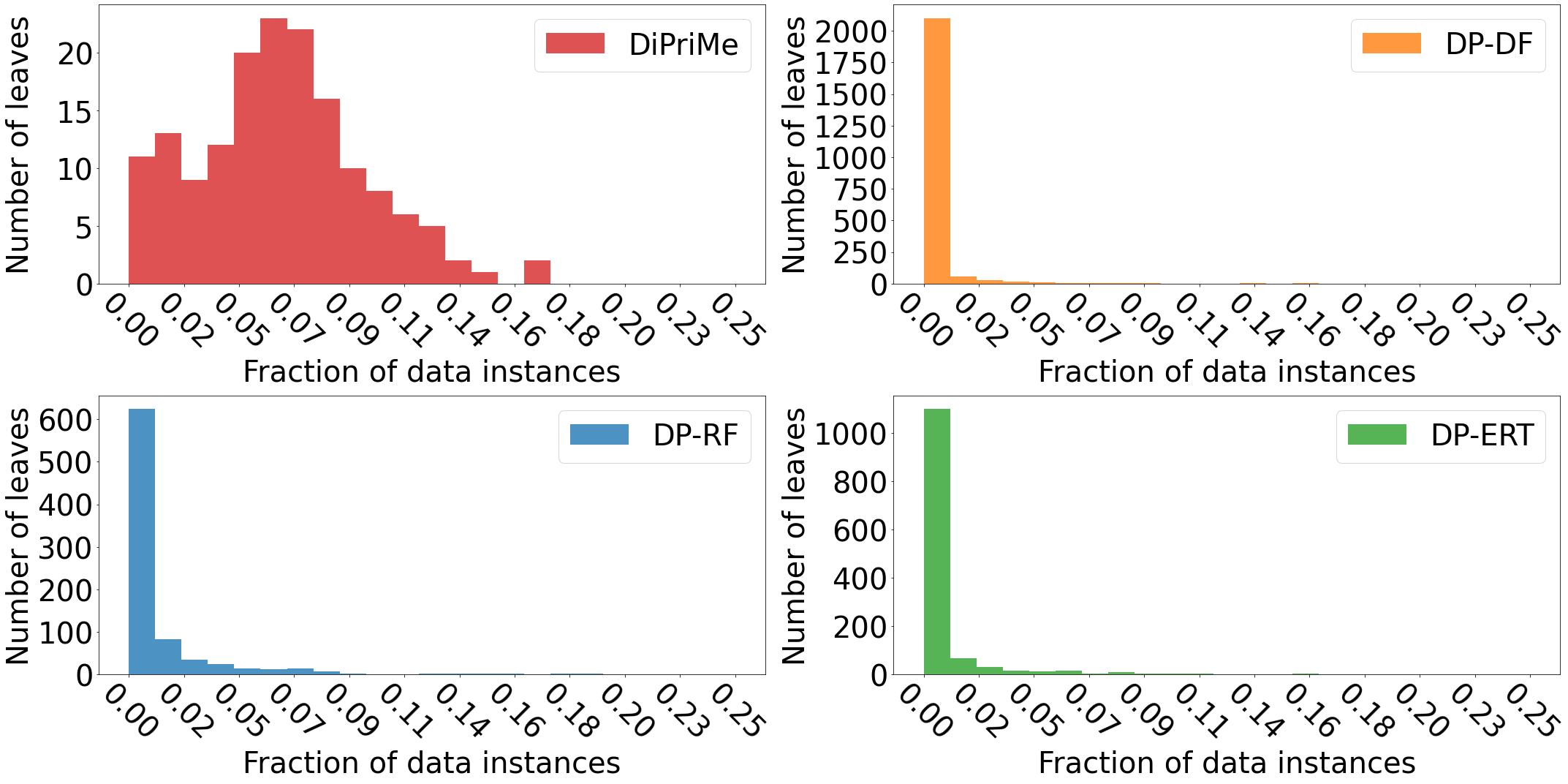}
    \caption{Fraction of data residing at each leaf node for Robot data.}
    \label{fig:node_occ}
\end{figure}


Our hypothesis for the superior performance of DiPriMe is that it discourages low occupancy nodes, which in turn reduces the impact of the injected noise on selecting optimal splits and estimating leaf-node parameters. Besides being borne out by Thms~\ref{thm:util_bound} and ~\ref{thm:class_bound}, a closer inspection of the tree ensembles learned on the Robot data evinces the claim that DiPriMe trees discourage low-occupancy nodes. Fig.~\ref{fig:node_occ} shows that DP-ERT, DP-RF and DP-DF have very heavy-tailed distributions over the per-leaf node occupancy, with the majority of nodes having very low occupancy. This is exacerbated in the trees generated by DP-RF and DP-DF, where each split generates multiple child nodes, one for each category of the selected attribute, compared with the binary splits used by DiPriMe and DP-ERT: DP-DF and DP-RF generated an average of 407 nodes and 104 nodes respectively, compared with 31 for DiPriMe and DP-ERT. In contrast, we see that the DiPriME trees have a much larger proportion of higher occupancy nodes.



\section{Discussion}\label{sec:disc}
DiPriMe is a new, differentially private, tree-based method for regression and classification. Rather than directly privatizing a successful non-private algorithm, DiPriMe explicitly aims to construct trees with balanced leaf node occupancy, exploiting the fact that privacy-inducing noise overwhelms the signal at low occupancy leaf nodes. This insight---which we back up with theoretical utility bounds---allows us to demonstrate impressive utility compared with competing methods, while maintaining the same level of privacy.

\bibliographystyle{icml2021}
\bibliography{refs}

\newpage
\onecolumn
\appendix
\raggedright
\section*{Appendix}
\newcommand{\varA}{\sigma_{\mathcal{A}}^2}
\newcommand{\varAj}[1]{\sigma_{\mathcal{A}|#1}^2}

We derive sensitivities used in the main paper in Section~\ref{sec:sensitivity} and provide proofs for the theorems in Section~\ref{sec:proofs}. Section~\ref{sec:add_results} provides some additional experimental results.


\section{Sensitivity analysis}\label{sec:sensitivity}
We assume have a set $\mathcal{A}$ with data points $\{y_1,y_2,\dots,y_N\}$ such that $|y_i| \leq B$. $\mathcal{A}\setminus j$ denotes the set of all data points besides $y_j$, i.e., $\mathcal{A}\setminus j = \{y_1,y_2,...y_{j-1},y_{j+1},...,y_N\}$. 
We consider the sensitivity $\Delta(f)$ of a statistic $f$ calculated on $\mathcal{A}$ to be the maximum change in $f$ due to adding or removing a single datapoint to $\mathcal{A}$. An alternative definition is the maximum change due to replacing a single datapoint in $\mathcal{A}$; in this case, the sensitivity is double the reported value.

\subsection{Mean}

\begin{equation*}
    \norm{\mu_{\mathcal{A}} -  \mu_{\mathcal{A}|j}} = \norm{\frac{y_j}{N}-\frac{1}{N(N-1)}\sum_{i=1,i\neq j}^N {y_i}} \leq \frac{2B}{N} 
\end{equation*}

$\therefore \Delta(\mu_{\mathcal{A}}) = \frac{2B}{N}.$

\subsection{Mean squared error}
The mean squared error is equivalent to the variance. Denoting the variance by $\sigma^2$,
\begin{align*}
\varA &= \frac{1}{N}\sum_{i=1}^N {y_i^2} -                 \frac{1}{N^2}\left(\sum_{i=1}^N{y_i}\right)^2\\
\end{align*}

Using the triangle inequality repeatedly, we get
\begin{align*}
\norm{\varA - \sigma^2_{\mathcal{A}\setminus j}} 
    &\leq \norm{\frac{N-1}{N^2}B^2} + \norm{\frac{N-1}{N^2}B^2}+\norm{ \frac{2(N-1)}{N^2}B^2} \\
    &\leq \frac{4B^2}{N} \\
\end{align*}
$\therefore \Delta(\varA) = \frac{4B^2}{N}.$

\section{Proofs}\label{sec:proofs}
Let $D = (X,Y)$ be a set of $N$ data points ordered according to an input attribute, so that $x_1\leq x_2\leq\cdots\leq x_N$.  Let $D_L, D_R$ be the subsets of $D$ obtained by splitting on the median value of $X$, and. Let $\widetilde{D}_L,\widetilde{D}_R$ be the subsets obtained by a random mechanism such that $Pr\left(\min\{|\widetilde{D}_L|,|\widetilde{D}_R|\})\leq t\right)=\zeta_t$ for $0<t<N/2$.

\subsection{Theorem~\ref{thm:util_bound}: Bounding the change in sum squared loss due to privatization of median tree regressor}
\textit{Let $SSE$ be the sum of squared errors under the split $D_L, D_R$ using exact means. Let $\widetilde{SSE}$ be the sum of squared errors under the split $\widetilde{D}_L,\widetilde{D}_R$ where the means have been rendered $\epsilon_\ell$-DP as described in Sec~\ref{sec:diprime}. Then for any $0<t<N/2$, $ \E[\widetilde{SSE}] - SSE \leq 4B^2N - 8B^2t + \frac{16B^2}{\epsilon_\ell^2 t}$ w.p.\ at least $1-\zeta_t$.}

\begin{proof}
Consider a non-private median split. There are $c=N/2$ points in each resulting block, i.e., $|D_L| = |D_R| = N/2$. W.l.o.g, let the target values of these data points be $\mathcal{A} = \{y_1,y_2,\dots,y_c\}$. Then, the value of our objective function (sum squared loss) in the block $D_L$ $a$ is 
\begin{align*}
    \mbox{SSE}_L &= \sum_{i=1}^{c} (y_i - \bar{y}_1)^2 \\
    &= c\varS{A}.
\end{align*}

Assuming $|\widetilde{D}_L| = c-\delta$, our noisy objective value is 
\begin{equation*}
    \mbox{SSE}_L^* = (c-\delta)\varS{C}.
\end{equation*}

\paragraph{Sensitivity of SSE} W.l.o.g., we take excluded points to be $\left\{y_1,y_2,\dots y_{\delta}\right\}$. We assume all the points are bounded, i.e., $|y_i| \leq B$. Then,

\begin{align*}
    c\varS{A} &= \sum_{i=1}^c {y_i^2} - \frac{1}{c}\left(\sum_{i=1}^c{y_i}\right)^2\\
    (c-\delta)\varS{B} &= \sum_{i=\delta+1}^c {y_i^2} - \frac{1}{c-\delta}\left(\sum_{i=\delta+1}^c{y_i}\right)^2.
\end{align*}

Hence,
\begin{align*}
    \norm{c\varS{A}-(c-\delta)\varS{B}} &= \norm{\frac{\delta}{c(c-\delta)}\left(\sum_{i=\delta+1}^c{y_i}\right)^2} \\
    & \qquad+ \norm{\sum_{i=1}^\delta{y_i^2}- \frac{1}{c}\left(\sum_{i=1}^\delta{y_i}\right)^2} \\& \qquad +  \norm{\frac{2}{c}\left(\sum_{i=1}^\delta{y_i}\right)\left(\sum_{i=\delta+1}^c y_i\right)} \\
    &\leq \frac{\delta(c-\delta)}{c}B^2 + \delta B^2 + \frac{2\delta(c-\delta)}{c}B^2 \\
    &\leq 4B^2 \delta. \\
\end{align*}

\paragraph{Returning to the utility analysis,} we see that noising the split by $\delta$ leads to a maximum change of $4B^2\delta$ in SSE at each leaf node, i.e.,
\begin{align}
\label{eq:sse_change_med}
\left\lvert|\widetilde{D}_L| - \frac{N}{2}\right\rvert
\leq t \implies |\mbox{SSE}_L^* - \mbox{SSE}_L| \leq 4B^2 t.
\end{align}

Let us now analyze the effect of privatizing the means as described in Sec~\ref{sec:diprime}. Denoting $|\widetilde{D}_L|$ as $N_L^*$, it is simple to observe that
\begin{align*}
    \widetilde{\mbox{SSE}}_L = \mbox{SSE}_L^* + N_L^*\rho_L^2
\end{align*}
where $\rho_L \sim \mbox{Laplace}(0, 2B/N_L^*\epsilon_\ell)$. \medskip

The conditional expectation of the perturbation due to this noise is
\begin{align*}
    \E[N_L^*\rho_L^2|N_L^*] = \frac{8B^2}{N_L^* \epsilon_\ell^2}.
\end{align*}

Note that $N_L^* \geq t \implies \E[N_L^*\rho_L^2|N_L^*] \leq  \frac{8B^2}{\epsilon_\ell^2 t}$. Combining the effects of the noise at both leaf nodes with \eqref{eq:sse_change_med}, we arrive at the desired quantity ($N_R^* = |\widetilde{D}_R|$)
\begin{align*}
    \E[\widetilde{SSE}] - SSE &= (SSE_L^* - SSE_L) + (SSE_R^*-SSE_R) + \E[N_L^*\rho_L^2] + \E[N_R^*\rho_R^2] \\
    &\leq 8B^2\left(\frac{N}{2}-t\right) + \frac{16B^2}{\epsilon_\ell^2 t}
\end{align*}
with probability at least $1-\zeta_t$ where $\zeta_t = Pr\left(\min\{|\widetilde{D}_L|,|\widetilde{D}_R|\})\leq t\right)$.
\end{proof}

\subsection{Theorem~\ref{thm:class_bound}: Bounding the change in accuracy due to privatization of median tree classifier}
\textit{Let $Acc$ be the classification accuracy under the split $D_L, D_R$ using exact counts. Let $\widetilde{Acc}$ be the accuracy under the split $\widetilde{D}_L,\widetilde{D}_R$ where the counts have been rendered $\epsilon_\ell$-DP as described in Sec~\ref{sec:diprime}. Then for any $0<t<N/2$, $\mbox{Acc} - \E[\widetilde{Acc}] \leq \left\{\frac{\epsilon_\ell t(2\theta_m-1)}{4} + \frac{1}{2}\right\}e^{-\epsilon_\ell t(2\theta_m-1)}$ 
w.p.\ at least $1-\zeta_t$ where $\theta_m$ is the minimum purity of a leaf node.
}

\begin{proof}
Without loss of generality, we shall number the classes as ``0" and ``1". We denote the respective class counts in a leaf node of the private median tree as $n_0^*$ and $n_1^*$. Privatization of these class counts gives class counts 
\begin{align*}
\tilde{n}_j \sim \max\{0, n_j + \mbox{Laplace}(0,1/\epsilon_\ell)\}, j\in\{0,1\}.
\end{align*}
Note that if the perturbed class counts are both zero, the prediction from that node is a randomly selected from $\{0,1\}$. \\
\vspace{1em}
Consider the case where $n_0^* < n_1^*$. The prediction at the leaf node will flip due to privatization if $\tilde{n}_0 > \tilde{n}_1$. Denoting this as the probability of flipping the prediction, $P_f$, we have
\begin{align*}
    P_f &= Pr(\tilde{n}_0 > \tilde{n}_1) \\
    &= \int_0^\infty{Pr(\tilde{n}_1 < x) f_{\tilde{n}_0}(x) dx} + \frac{1}{2}Pr(\tilde{n}_0 = 0, \tilde{n}_1 = 0) \\
    &= \int_0^{n_0^*}\left\{\frac{1}{2} e^{{\epsilon_\ell}(x-n_1^*)}\right\}\left\{\frac{\epsilon_\ell}{2}e^{{\epsilon_\ell}(x-n_0^*)}\right\}dx \\
    &+ \int_{n_0^*}^{n_1^*}\left\{\frac{1}{2} e^{{\epsilon_\ell}(x-n_1^*)}\right\}\left\{\frac{\epsilon_\ell}{2}e^{{\epsilon_\ell}(n_0^*-x)}\right\}dx \\
    &+ \int_{n_1^*}^\infty\left\{1-\frac{1}{2} e^{{\epsilon_\ell}(n_1^*-x)}\right\}\left\{\frac{\epsilon_\ell}{2}e^{{\epsilon_\ell}(n_0^*-x)}\right\}dx \\ &+  \frac{1}{2}\left(\frac{1}{2} e^{{-\epsilon_\ell}n_0^*}\right)\left(\frac{1}{2} e^{{-\epsilon_\ell}n_1^*}\right) \\
    &= \left\{\frac{\epsilon_\ell(n_1^*-n_0^*)}{4} + \frac{1}{2}\right\}e^{\epsilon_\ell(n_0^*-n_1^*)}
\end{align*}

By symmetry, we can generalize $P_f$ as 
\begin{align*}
    P_f &= \left\{\frac{\epsilon_\ell|n_1^*-n_0^*|}{4} + \frac{1}{2}\right\}e^{-\epsilon_\ell|n_1^*-n_0^*|} \\
\end{align*}

Recall that the count in one of the resulting blocks is $N_L^* = |\widetilde{D}_L| = n_0^*+n_1^*$. Hence, we can rewrite the above expression for this block as
\begin{align*}
    P_{f,L} &= \left\{\frac{\epsilon_\ell N_L^*(2\theta_L^*-1)}{4} + \frac{1}{2}\right\}e^{-\epsilon_\ell N_L^*(2\theta_L^*-1)} 
\end{align*}
where $\theta_L^*$ is the purity of the block, i.e., fraction of dominant class in the block $\widetilde{D}_L$. Note that $P_{f,L}$ is monotonically decreasing over $N_L^*$ and $\theta_L^*$.

Following along the lines of the proof for Theorem \ref{thm:util_bound}, $P_{f,L}\leq \left\{\frac{\epsilon_\ell t(2\theta_L^*-1)}{4} + \frac{1}{2}\right\}e^{-\epsilon_\ell t(2\theta_L^*-1)}$ if $N_L^* \geq t$.

Accumulating the effect of privatization on both resulting blocks, 
\begin{align*}
    Acc - \E[\widetilde{Acc}] &\leq \frac{1}{N}\sum_{j=L,R} N_j^*\left\{\frac{\epsilon_\ell t(2\theta_j^*-1)}{4} + \frac{1}{2}\right\}e^{-\epsilon_\ell t(2\theta_j^*-1)} \\
    &\leq \left\{\frac{\epsilon_\ell t(2\theta_m-1)}{4} + \frac{1}{2}\right\}e^{-\epsilon_\ell t(2\theta_m-1)}
\end{align*}
with probability at least $1-\zeta_t$ and each leaf node has a minimum purity $\theta_m$. \\
\vspace{1em}
The additional parameter $\theta_m$ intuitively makes sense as impure leaf nodes, i.e, $\theta_i  \approx 1/2$ are likely to flip predictions upon privatization.
\end{proof}

\subsection{Theorem \ref{thm:occupancy}: Median splits discourage imbalanced splits given enough spread in the data}
\textit{Consider $N$ data points in $[x_L, x_U]$, such that $R=x_U-x_L$, where the $N-2t$ centermost data points cover a range of length at least $d<R$. If we use EM to generate an $\epsilon_s$-DP median split, then the smaller of the two resulting blocks will contain at least $t$ data points w.p.\ at least $\frac{d}{Re^{-\epsilon_s}+d(1-e^{-\epsilon_s})}$.}

\begin{proof}
W.l.o.g., we can assume that the feature values of the data points in ascending order are $\mathcal{A} = \{x_1, x_2, \dots, x_N\}$. We denote the size of the smaller of the two blocks resulting from the private median split result as $N_c^{med}$. \\
\medskip
Recall that the scoring function  $q(r) = -\lvert|\mathcal{A} \intersect [a_L, r)| - |\mathcal{A} \intersect [r, a_U]|\rvert$. This is piecewise constant between successive data points, i.e., $q(r)$ is constant for any $r\in (x_i, x_{i+1})$. We shall denote this value by $q_i$. Clearly, $q_i = q_{N-i}$ and $q_{i+1} = q_i+2$ for $i<N/2$. Then, we have 
\begin{align}
    Pr(N^{med}_c \leq t) &\leq \frac{(R-d)e^{\epsilon_s q_t/2}}{(R-d)e^{\epsilon_s q_t/2} + de^{\epsilon_s q_{t+1}/2}} \nonumber\\
    &= \frac{(R-d)e^{-\epsilon_s}}{(R-d)e^{-\epsilon_s} + d}.
\end{align}
The stated result is the probability of the complement of the event.
\end{proof}

\subsubsection{Corollary~\ref{cor:rand}: Median splits are better than random splits}
\textit{Consider $N$ data points in some bounded space. Let $N^{med}_c$ be the size of the smaller of the two resulting blocks upon splitting at the differentially private median, and $N^{rand}_c$ be the corresponding number for a randomly selected split. Then, for any $0 < t < N/2$, $Pr(N^{med}_c\leq t) \leq Pr(N^{rand}_c\leq t)$.
}
\begin{proof}

It is straightforward to see that $Pr(N^{rand}_c\leq t) = \frac{R-d}{R}$.
\begin{align*}
    Pr(N^{med}_c\leq t) - Pr(N^{rand}_c\leq t) &\leq \frac{(R-d)e^{-\epsilon}}{(R-d)e^{-\epsilon} + d} - \frac{R-d}{R} \\
    &= \frac{d}{R} - \frac{d}{(R-d)e^{-\epsilon} + d} \\
    &\leq 0
\end{align*}
\end{proof}

\begin{figure}[H]
    \centering
        \includegraphics[width=.45\columnwidth]{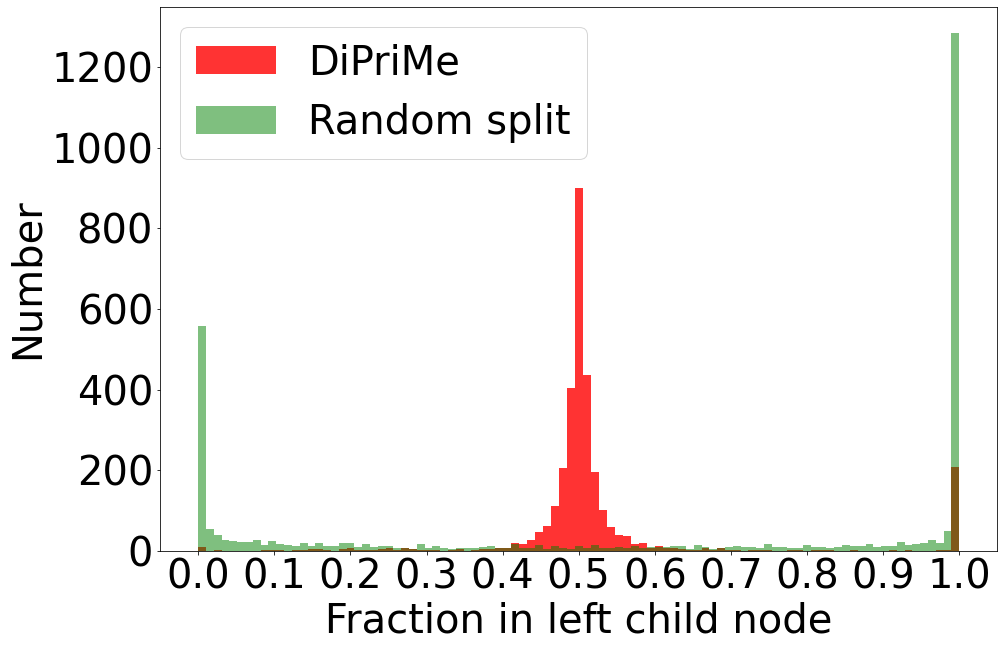}
        \caption{Fraction of instances assigned to left child, on Banknote data.}
        \label{fig:hist_frac}
\end{figure}

We inspect the fraction of data points assigned to the the left child of each node, i.e., $|D_L|/|D|$, for DiPriMe trees and DP-ERT learned on the Banknote Authentication dataset in Figure~\ref{fig:hist_frac}. This corroborates the result of Corollary~\ref{cor:rand}, that private median splits are more balanced than random splits.

\section{Additional results}\label{sec:add_results}
\begin{figure}[H]
    \centering
    \begin{subfigure}[t]{0.32\textwidth}
        \includegraphics[width=\textwidth]{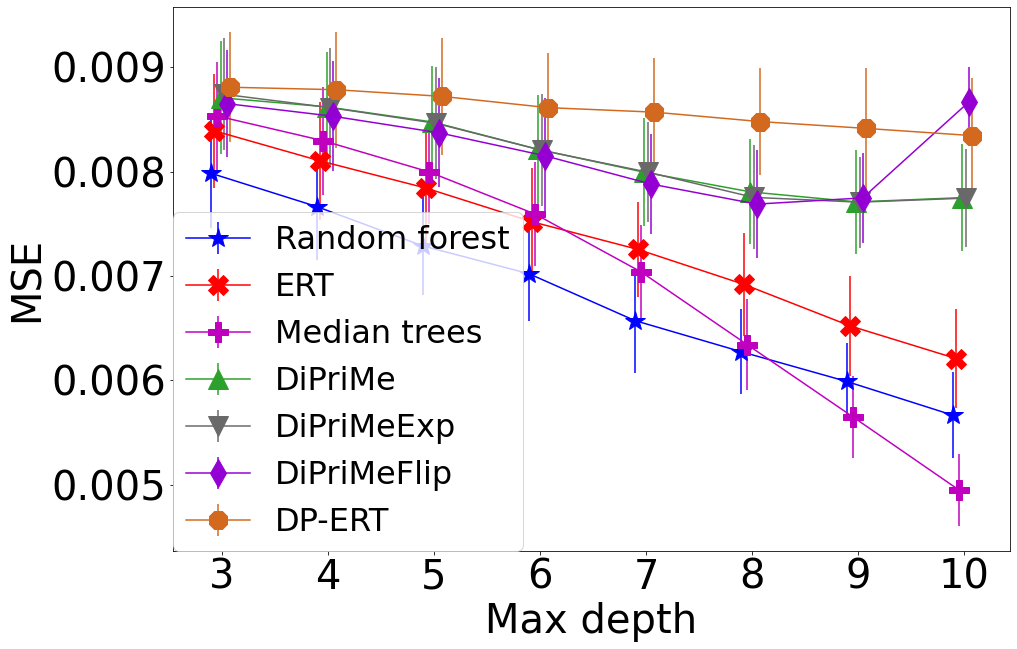}
        \caption{$\epsilon=10$, $N_T=10$}
        \label{fig:reg_maxd_all}
    \end{subfigure}
    \centering
    \begin{subfigure}[t]{0.32\textwidth}
        \includegraphics[width=\textwidth]{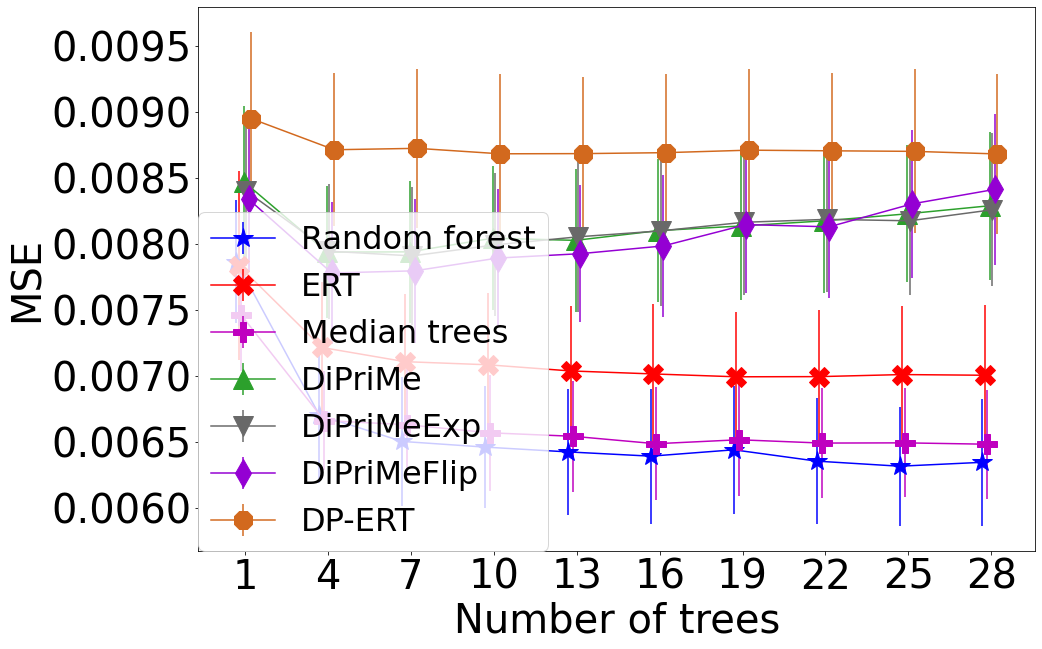}
        \caption{$\epsilon=10$, $d_{max}=8$}
        \label{fig:reg_ntree_all}
    \end{subfigure}
    \centering
    \begin{subfigure}[t]{0.32\textwidth}
        \includegraphics[width=\textwidth]{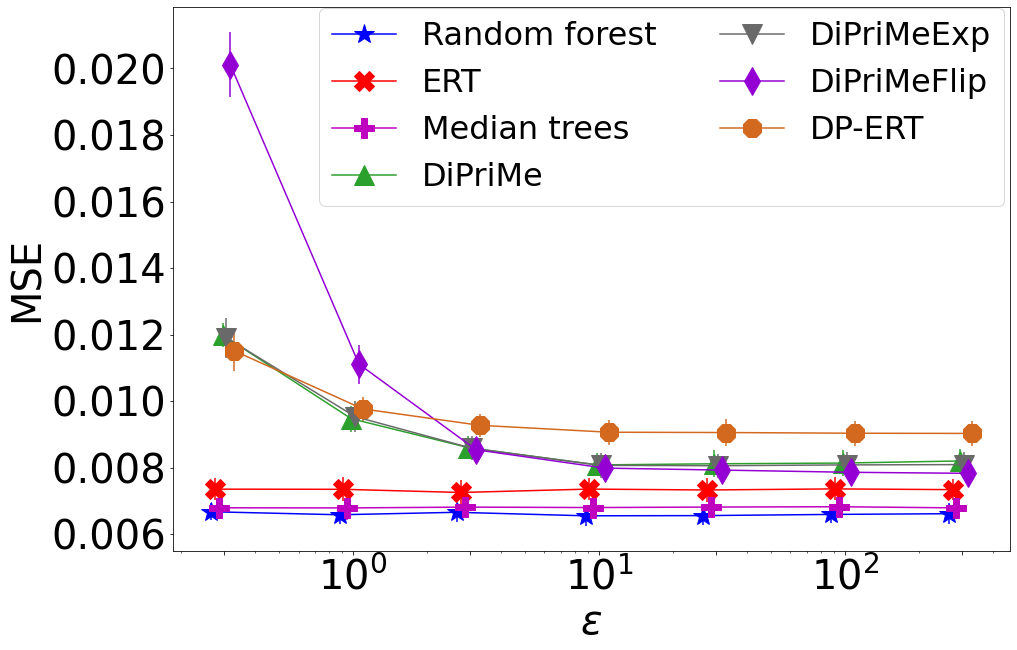}
        \caption{$d_{max}=8$, $N_T=10$}
        \label{fig:reg_eps_all}
    \end{subfigure}
    \caption{Mean squared error of DiPriMe without partitioning data, Random Forest, Extremely 
    Randomized Trees and DP-ERT at various values of $\epsilon$, $d_{max}$ and $N_T$ on the Appliances Energy prediction dataset.}
    \label{fig:reg_param_all_data}
\end{figure}

\begin{figure}[h]
    \centering
    \includegraphics[width=0.4\textwidth]{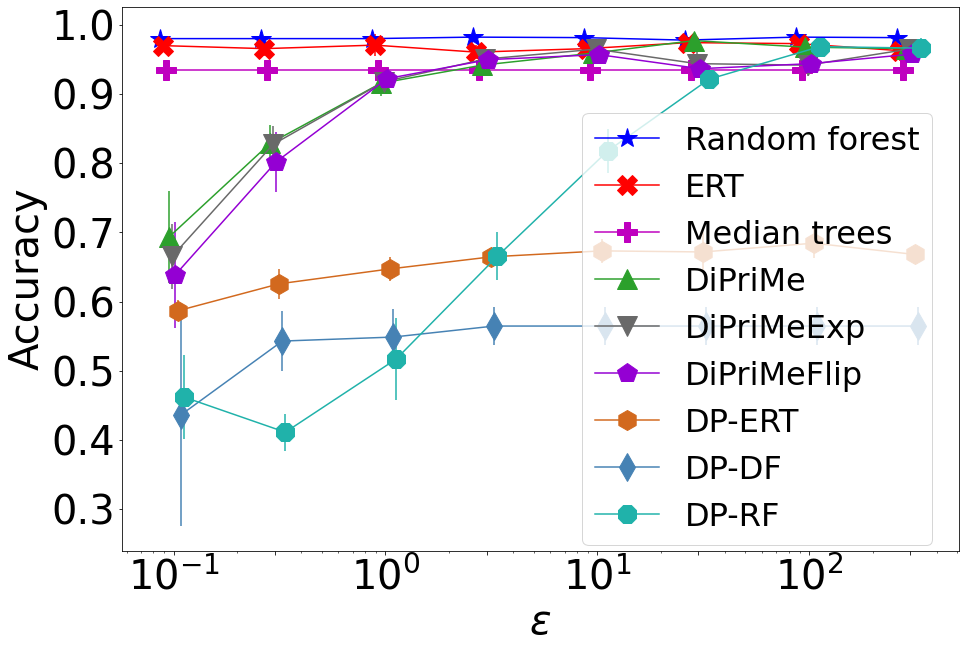}
        \caption{Performance of DiPriMe without partitioning data, Random Forest, Extremely Randomized Trees, DP-ERT and DP-DF at various values of $\epsilon$ for the Banknote Authentication data ($d_{max} = 5$, $N_T=10$, $\rho = 0.5$).}
    \label{fig:class_eps_all}
\end{figure}

Figure \ref{fig:reg_param_all_data} displays the results for DiPriMe and its variants without paritioning the data for each tree. The trends are similar to those seen in Figure \ref{fig:reg_param}. An observation of note is that the optimal maximum depth for trees fit on all the data is higher than that fit on disjoint subsets of data. This is congruent with the intuition that low-occupancy nodes are noised more heavily. Hence, trees fitted to more data can be grown deeper before suffering from a similar loss of utility. This line of reasoning leads us to believe that learning an ensemble of DiPriME trees on disjoint subsets of data will be a more powerful learner with larger amounts of training data. We see a similar trend in Figure~\ref{fig:class_eps_all} to that in Figure~\ref{fig:class_eps} for the task of classification.
\vskip 0.1in

\begin{figure}[ht]
    \centering
    \begin{subfigure}[t]{0.45\textwidth}
    \includegraphics[width=\textwidth]{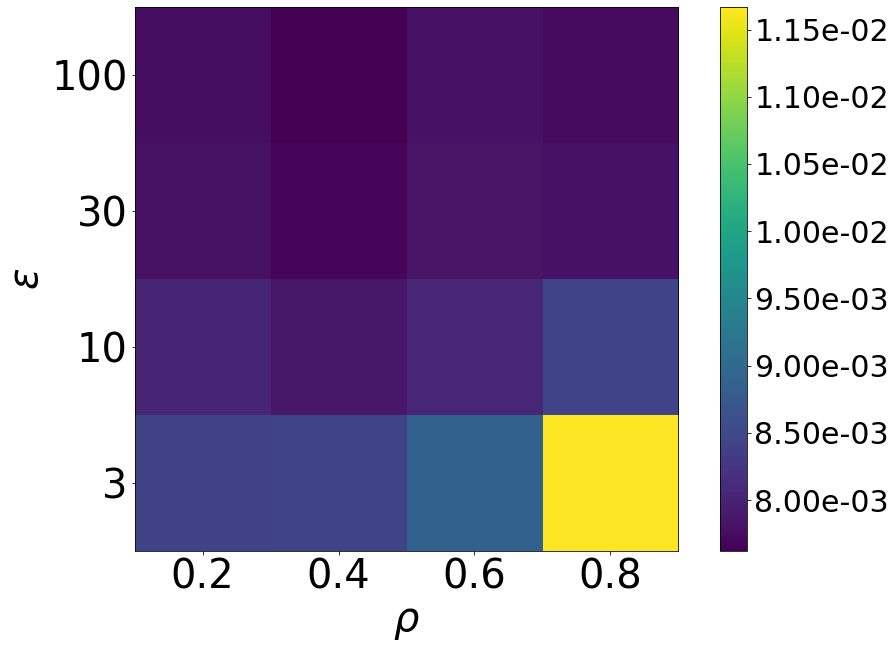}
    \label{fig:dpe_hmap_eps_rho}
    \caption{$N_T=10$, $d_{max}=8$, $K=10$, \\ data partitioned}
    \vspace{1em}
    \end{subfigure}
    \centering
    \begin{subfigure}[t]{0.45\textwidth}
    \includegraphics[width=\textwidth]{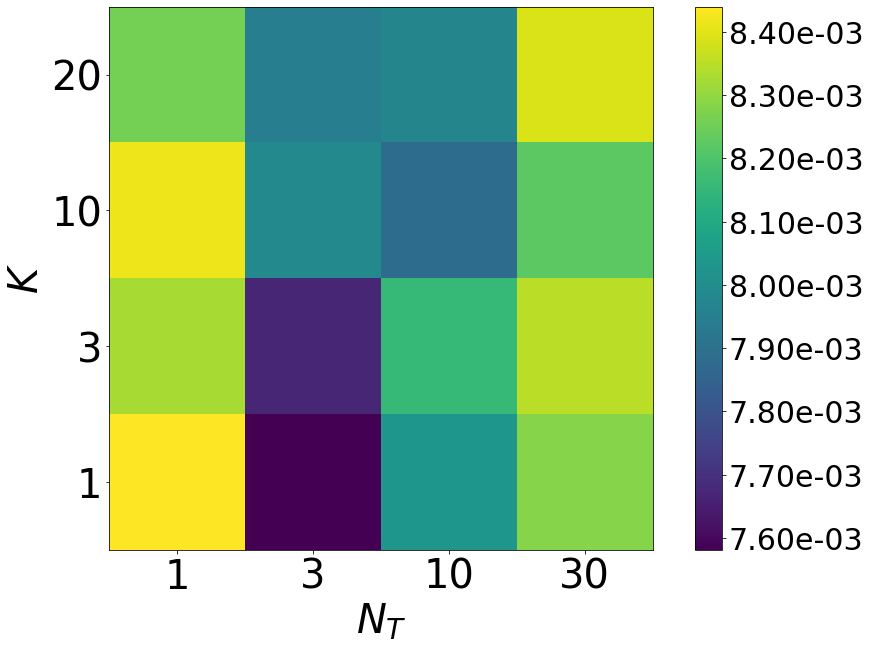}
    \label{fig:dpe_hmap_NT_K}
    \caption{$d_{max}=8$, $\epsilon=10$, $\rho = 0.5$, \\ data partitioned}
    \vspace{1em}
    \end{subfigure}
    \centering
    \begin{subfigure}[t]{0.45\textwidth}
    \includegraphics[width=\textwidth]{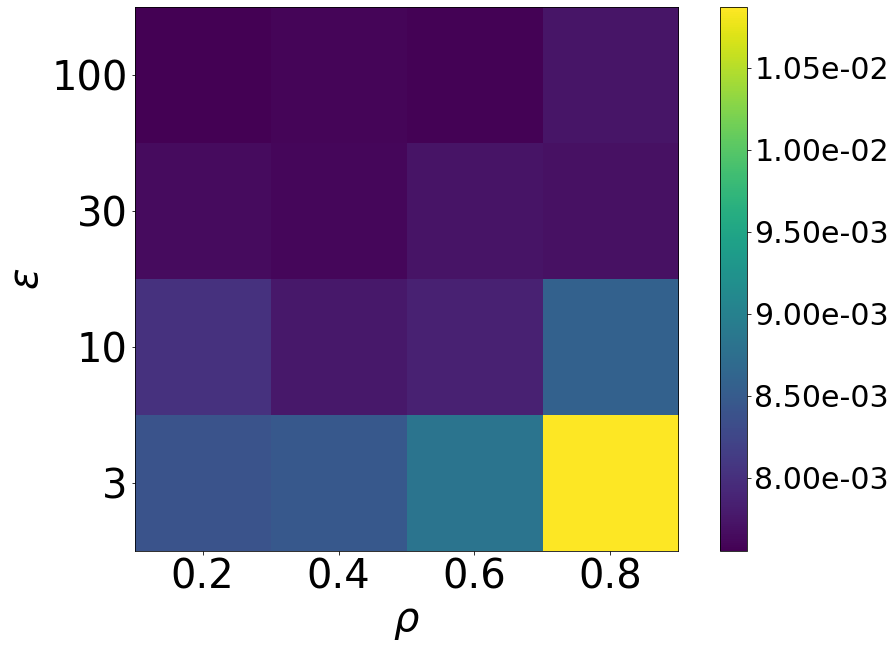}
    \label{fig:dpe_hmap_eps_rho_all_data}
    \caption{$N_T=10$, $d_{max}=8$, $K=10$, \\ data not partitioned}
    \end{subfigure}
    \centering
    \begin{subfigure}[t]{0.45\textwidth}
    \includegraphics[width=\textwidth]{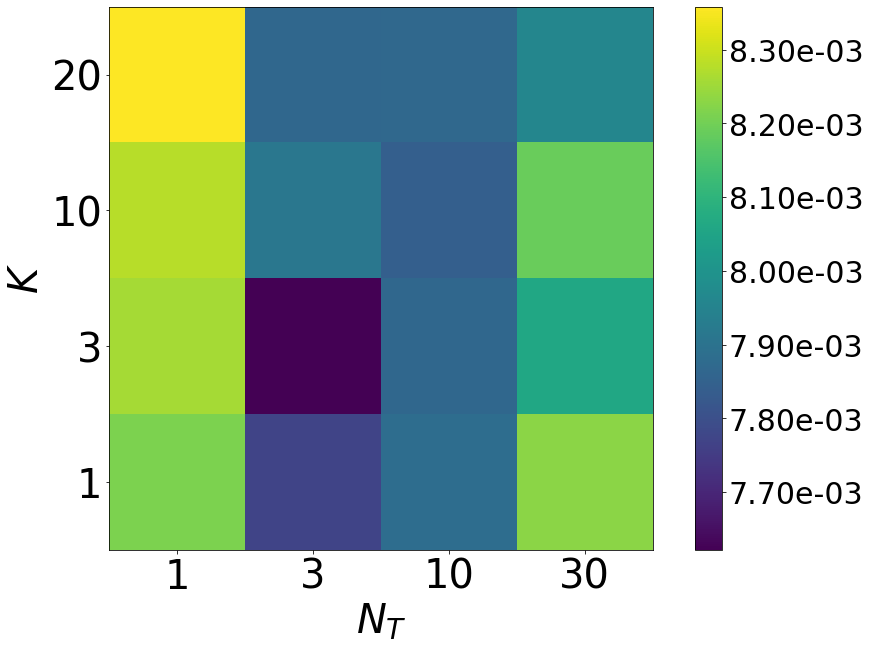}
    \label{fig:dpe_hmap_NT_K_all_data}
    \caption{$d_{max}=8$, $\epsilon=10$, $\rho = 0.5$, \\ data not partitioned}
    \end{subfigure}
    \caption{Mean squared error of DiPriMeExp on the Appliances Energy Prediction dataset for various hyperparameter settings}
    \label{fig:heatmap_DPE}
\end{figure}

\begin{figure}[h]
    \centering
    \begin{subfigure}[t]{0.45\textwidth}
    \includegraphics[width=\textwidth]{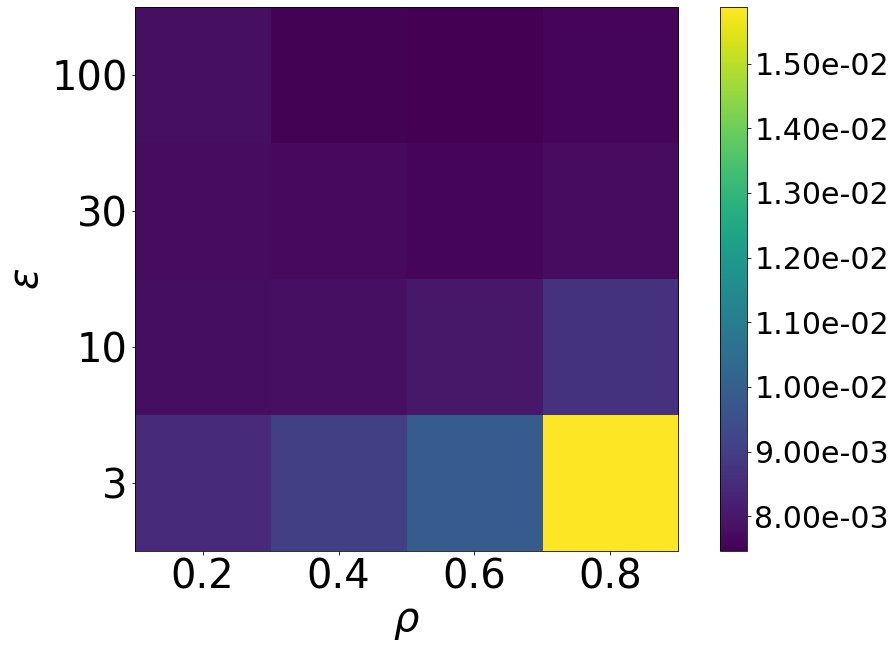}
    \label{fig:dpf_hmap_eps_rho}
    \subcaption{$N_T=10$, $d_{max}=8$, $K=10$, \\ data partitioned}
    \vspace{1em}
    \end{subfigure}
    \centering
    \begin{subfigure}[t]{0.45\textwidth}
    \includegraphics[width=\textwidth]{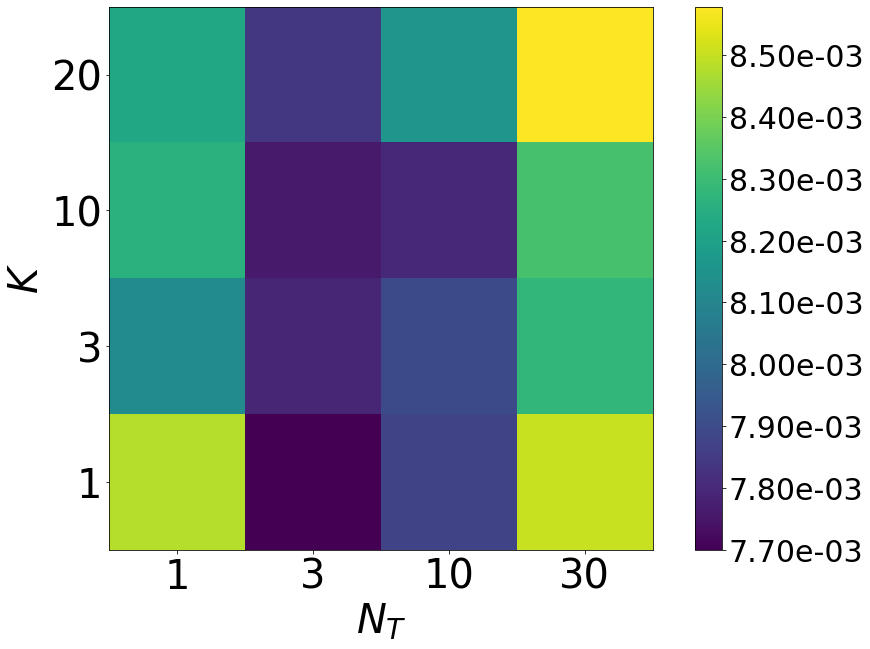}
    \label{fig:dpf_hmap_NT_K}
    \subcaption{$d_{max}=8$, $\epsilon=10$, $\rho = 0.5$, \\ data partitioned}
    \vspace{1em}
    \end{subfigure}
    \centering
    \begin{subfigure}[t]{0.45\textwidth}
    \includegraphics[width=\textwidth]{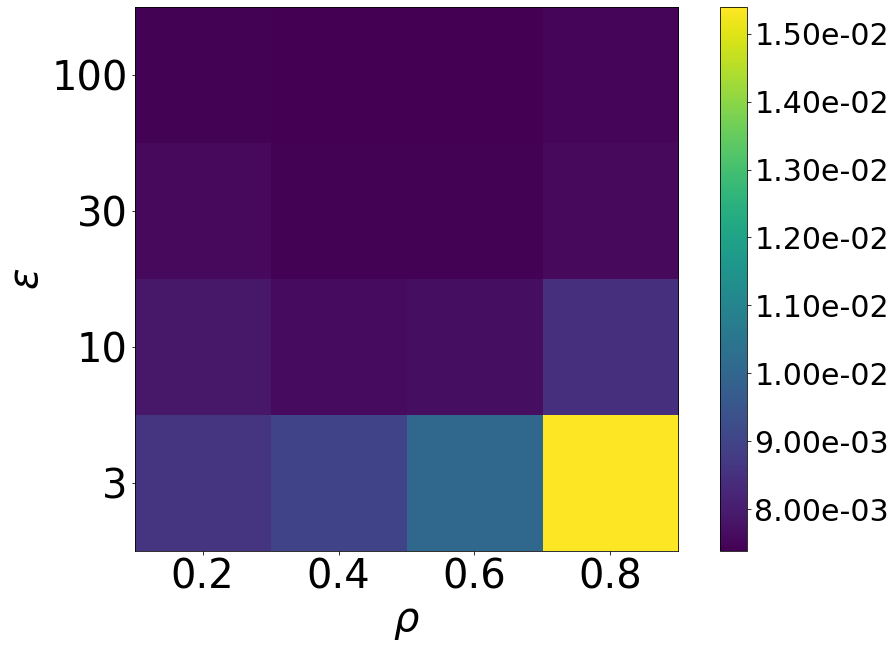}
    \label{fig:dpf_hmap_eps_rho_all_data}
    \subcaption{$N_T=10$, $d_{max}=8$, $K=10$, \\ data not partitioned}
    \end{subfigure}
    \centering
    \begin{subfigure}[t]{0.45\textwidth}
    \includegraphics[width=\textwidth]{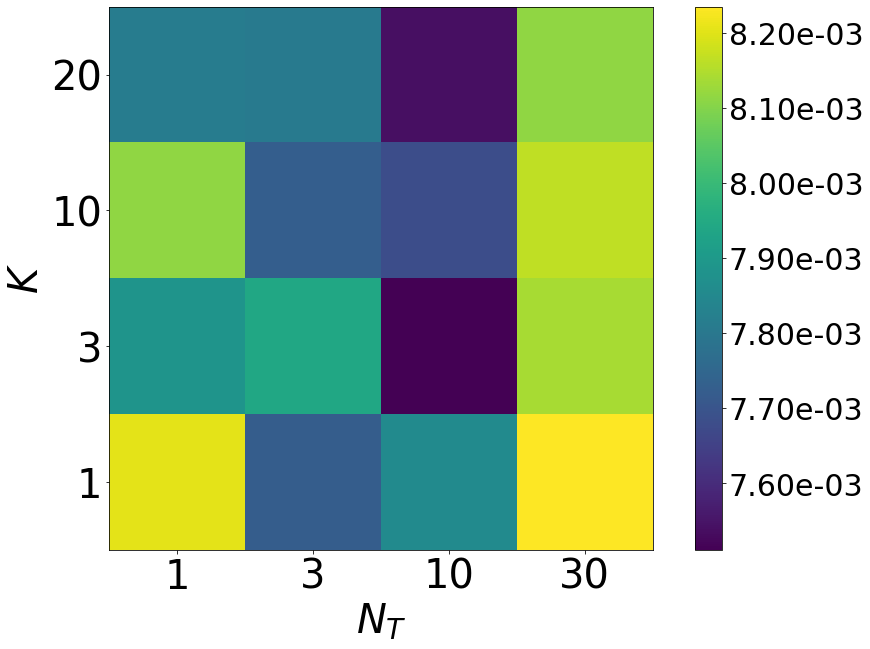}
    \label{fig:dpf_hmap_NT_K_all_data}
    \subcaption{$d_{max}=8$, $\epsilon=10$, $\rho = 0.5$, \\ data not partitioned}
    \end{subfigure}
    \caption{Mean squared error of DiPriMeFlip on the Appliances Energy Prediction dataset for various hyperparameter settings}
    \label{fig:heatmap_DPF}
\end{figure}

Figures \ref{fig:heatmap_DPE} and \ref{fig:heatmap_DPF} once again exhibit the improved performance of DiPriMeExp and DiPriMeFlip with larger privacy budgets. It also shows that increasing $N_T$ improves performance only to a certain limit before the increased noise reduces the utility of the DiPriMe trees. The key insight here is the importance of the hyperparameter $\rho$ for good performance; large values of $\rho$ leaves less privacy budget for storing the means, resulting in deterioration in the MSE. This effect is more pronounced at smaller values of $\epsilon$ as the noise scales as $1/\epsilon$. Note that data-driven hyperparameter selection has to be done privately as well \cite{liu2019private}. This would require additional privacy budget, i.e., a larger $\epsilon$.

\end{document}